%% file: main.tex
\documentclass[11pt,a4paper]{article}

\usepackage{titlesec}
\usepackage{authblk}

\usepackage[utf8]{inputenc}
\usepackage[english]{babel}
\usepackage[T1]{fontenc} % Font encoding
\usepackage{lmodern} % Latin Modern (scalable Computer Modern)
\usepackage{microtype} % Character protrusion and font expansion for better justification
\usepackage{csquotes}

\usepackage{graphicx}
\usepackage{caption}
\usepackage{subcaption}

\usepackage{amsmath}
\usepackage{bm}
\usepackage[overload]{empheq}
\usepackage{amsfonts}

\usepackage{multirow}
\usepackage{booktabs}
\usepackage{makecell}
\usepackage{threeparttable}

\usepackage{algorithm}
\usepackage{algpseudocode}

\usepackage[colorlinks=true,linkcolor=black,anchorcolor=black,citecolor=black,filecolor=black,menucolor=black,runcolor=black,urlcolor=black]{hyperref} % Adds clickable links at references
\usepackage[sorting=none, style=numeric-comp]{biblatex}
\DeclareSourcemap{
  \maps[datatype=bibtex]{
    \map{
      \step[fieldset=urldate, null]
    }
  }
}
\appto{\bibsetup}{\sloppy}

\usepackage{xcolor}
\usepackage{tikz}
\usepackage{tikz-3dplot}
\usetikzlibrary{arrows.meta, patterns}

\usepackage{geometry}
\input{macros}

\title{
Learning PDE solution operators with variable initial conditions via Latent Dynamics Networks
}

\author[1]{\small Stefano Maria Pizzamiglio}
\author[1]{\small Stefano Pagani}
\author[1]{\small Francesco Regazzoni}
\affil[1]{\footnotesize MOX Laboratory, Department of Mathematics, Politecnico di Milano, Italy}
\affil[ ]{\footnotesize \texttt{\{\href{mailto:stefanomaria.pizzamiglio@polimi.it}{stefanomaria.pizzamiglio},
    \href{mailto:stefano.pagani@polimi.it}{stefano.pagani},
    \href{mailto:francesco.regazzoni@polimi.it}{francesco.regazzoni}\}@polimi.it}
    }

\date{}

\begin{document}

    \maketitle

    \begin{abstract}
        \input{Sections/abstract}
    \end{abstract}

    \section{Introduction}
    \label{sec:introduction}
    \input{Sections/introduction}
    
    \section{Methods}
    \label{sec:methods}
    \input{Sections/methods}

    \section{Results}
    \label{sec:results}
    \input{Sections/results}

    \section{Conclusions}
    \label{sec:conclusions}
    \input{Sections/conclusions}

    \section*{Acknowledgements}
    \noindent The present research has received support from the project FIS, MUR, Italy 2025-2028, Project code:
    FIS-2023-02228, CUP: D53C24005440001, ``SYNERGIZE: Synergizing Numerical Methods and Machine Learning for a
    new generation of computational models''.
    We acknowledge ISCRA for awarding this project access to the LEONARDO supercomputer,
    owned by the EuroHPC Joint Undertaking, hosted by CINECA (Italy).
    We acknowledge the grant Dipartimento di Eccellenza 2023-2027, MUR, Italy.
    We acknowledge ``INdAM - GNCS Project'' code CUP E53C25002010001.
    The authors of this work are members of GNCS, ``Gruppo Nazionale per il Calcolo Scientifico'' (National Group for Scientific Computing) of INdAM (Istituto Nazionale di Alta Matematica), Italy.

    \printbibliography

    \appendix
    \captionsetup{margin=0pt}

    \clearpage
    
    \section{Architecture details and hyperparameters}
    \label{sec:appendix-hyper}
    \input{Sections/appendix_hyperparameters}

    \clearpage
    
    \section{Training and validation}
    \label{sec:appendix-train-val}
    \input{Sections/appendix_training_validation}

\end{document}

%% file: macros.tex
\newcommand{\avsum}{\mathop{\mathpalette\avsuminner\relax}\displaylimits}

\makeatletter
\newcommand\avsuminner[2]{%
  {\sbox0{$\m@th#1\sum$}%
   \vphantom{\usebox0}%
   \ooalign{%
     \hidewidth
     \smash{\vrule height\dimexpr\ht0+1pt\relax depth\dimexpr\dp0+1pt\relax}%
     \hidewidth\cr
     $\m@th#1\sum$\cr
   }%
  }%
}
\makeatother

\newcommand{\argmin}{\operatornamewithlimits{argmin}}

%% file: Sections/abstract.tex
In many-query scenarios, data-driven surrogate models provide an efficient alternative to high-fidelity solvers for simulating physical systems governed by Partial Differential Equations (PDEs).
In this context, the Latent Dynamics Network (LDNet) has recently demonstrated remarkable performance in
predicting the response of spatio-temporal systems, combining Neural Ordinary Differential Equations with nonlinear dimensionality reduction.
However, the original formulation assumes a fixed initial condition, limiting its applicability to many real-world applications where a system evolves from varying starting states.
In this work, we overcome this limitation while keeping the end-to-end training procedure of the original LDNet and its encoder-free nature, which preserves its intrinsic independence from spatial resolution and grid topology.
We infer the initial latent state directly from a small set of early-time observations, treating latent-state initialization as an adaptation problem, and investigate two strategies: an auto-decoding formulation and a meta-learning approach in which the initial latent state acts as a task-specific context variable.
We demonstrate the accuracy of the proposed methods across diverse physical phenomena, spanning advection-diffusion, fluid dynamics, and solid mechanics.
Meta-learning markedly accelerates latent-state inference and induces smoother, better-conditioned optimization landscapes, and spontaneously organizes the latent space into a structured representation that reflects physically meaningful features of the underlying dynamics.
The coordinate-based decoder enables training from spatially subsampled data while recovering high-resolution solution fields at inference. The resulting approach provides an efficient and resolution-independent surrogate modeling framework for many-query simulations of time-dependent PDEs with varying initial conditions.

%% file: Sections/introduction.tex
Partial differential equations (PDEs) are fundamental for modeling complex physical systems
across various scientific and engineering domains.
However, in most applications, obtaining an analytical solution is intractable, necessitating the use of numerical
methods to compute approximate solutions~\cite{quarteroni_numerical_2014}.
Although high-fidelity methods, such as finite difference~\cite{leveque_finite_2007}, finite element~\cite{bathe_finite_1996}, or finite volume method~\cite{eymard_finite_2000}, can provide accurate results, they can be computationally expensive, especially in many-query settings (e.g., inverse design) or
for real-time predictions (e.g., control)~\cite{quarteroni_reduced_2016,hesthaven_certified_2016}.
Recently, data-driven approaches have emerged as a
promising alternative to address these challenges, leveraging the latest advancements in machine learning algorithms and High Performance Computing (HPC).
In particular, machine learning models can be trained to approximate the input-output behavior of high-fidelity numerical solvers, providing computationally efficient surrogates~\cite{thuerey_deep_2020}. These approaches fall within the broader field of Scientific Machine Learning (SciML), which develops learning-based methods for problems arising in scientific computing~\cite{karniadakis_physics-informed_2021,quarteroni_combining_2025}.

Traditionally, machine learning algorithms have been applied to map finite-dimensional inputs to finite-dimensional outputs. 
However, differential
problems
are characterized by inputs and outputs that belong to infinite-dimensional function spaces.
Specifically, given two function
spaces $\mathcal{U}$ and $\mathcal{V}$, an operator (e.g., the data-to-solution map associated with a PDE problem) $\mathcal{G}:\mathcal{U}\to\mathcal{V}$ maps a function
$u\in\mathcal{U}$ to a
function $v\in\mathcal{V}$.
The goal of \emph{operator learning} is therefore to approximate the operator $\mathcal{G}$
with $\tilde{\mathcal{G}}_{\boldsymbol{\theta}}$,
parameterized by a set of learnable parameters $\boldsymbol{\theta}$~\cite{quarteroni_combining_2025}.
For example, the
seminal work on Deep Operator Networks (DeepONets) employs two deep neural networks (i.e., \emph{branch} and
\emph{trunk} networks) to approximate the operator $\mathcal{G}$~\cite{lu_learning_2021}.
More recently, Neural Operators (NOs),
characterized by a sequence of integral kernel layers, generalize the Feed-Forward Neural Network (FFNN)
architecture to infinite-dimensional spaces~\cite{kovachki_neural_2023,quarteroni_combining_2025}.
Various NOs have been proposed in the literature, with the
Fourier Neural Operator (FNO) and its variants being among the most widely used, leveraging the
Fast Fourier Transform (FFT) to efficiently compute
the integrals in the frequency domain~\cite{li_fourier_2021,li_fourier_2023,li_physics-informed_2024}.

Due to their
discretization
invariance, Neural
Fields (NFs) have
been demonstrated to be effective for operator learning~\cite{serrano_operator_2023, wang_cvit_2025}. Specifically, a neural field, also known as a coordinate-based neural network or an implicit neural representation, consists of a
neural network that approximates a continuous field (e.g., scalar, vector, or tensor field) by mapping
input coordinates to field values~\cite{xie_neural_2022}. In addition to their popularity in visual
computing applications (e.g., Neural Radiance Fields for 3D scene representation~\cite{mildenhall_nerf_2021}), 
their ability to continuously represent functions makes them natural candidates to approximate inputs and
solutions of differential problems in a discretization-invariant manner. Thus, many SciML applications have adopted the
neural field framework, beginning with physics-informed approaches such as
Physics-Informed Neural Networks (PINNs)~\cite{raissi_physics-informed_2019} or the
Deep Ritz Method (DRM)~\cite{e_deep_2018}.
Alternatively, in the context of operator learning for dynamics problems,
the Dynamics-aware Implicit Neural Representations (DINo) framework~\cite{yin_continuous_2023} introduced a neural-field-based
encode-process-decode
pipeline, which has been recently extended
in other works~\cite{serrano_operator_2023,serrano_infinity_2023, wen_reduced-order_2023, serrano_aroma_2024,
    wang_gridmix_2025, catalani_neural_2024, catalani_geometry_2025}.

The Latent Dynamics Network (LDNet)~\cite{regazzoni_learning_2024} is a deep learning architecture
designed to
learn the solution operator of time-dependent PDEs. It can be seen as a process-decode neural field, where two fully
connected neural networks (FCNNs) are used as a latent-dynamics processor and a decoder, respectively.
Specifically, the physical state is represented by a compact latent vector, and the network is able to predict the value of the PDE solution at any time and at any spatial position within the PDE domain.
The LDNet has been successfully
applied to the surrogate modeling of several time-dependent PDEs, ranging from advection-diffusion-reaction problems
to computational fluid dynamics (CFD)
and cardiac electrophysiology~\cite{salvador_liquid_2026, centofanti_neural_2026,si_toward_2026,tomada_latent_2026}. Compared to other similar neural-field-based approaches, such as DINo or CORAL~\cite{serrano_operator_2023}, the LDNet learns the latent dynamics and the mapping back to the physical space simultaneously, in an end-to-end fashion. Training the dynamics and the decoder simultaneously not only simplifies the procedure, but also allows the neural surrogate to learn a latent representation that is optimal for both efficiently compressing the physical state and correctly predicting the evolution in time. 
However, the original LDNet implementation is restricted to problems starting
from a fixed initial condition.
This hinders its applicability in real-world scenarios, where the goal is to
predict the system evolution from an unseen initial state.
Notably, a concurrent graph-based extension of LDNet~\cite{tomada_latent_2026} explicitly identifies this as an open limitation of encoder-free latent-dynamics architectures, while addressing a complementary limitation (i.e., the lack of geometric inductive bias in the decoder).
Related architectures have tackled this limitation through different strategies: an explicit encoder~\cite{lin_parametric_2024, liang_parameterized_2026}, or a shifted time convention in which the dynamics network itself bootstraps the latent state from the initial observation~\cite{xiao_ld-ensf_2026,si_levda_2026,si_toward_2026}. Unlike these works, our approach remains entirely encoder-free and resolution-independent, and it does not require a specific time-stepping convention.
Encoder-free latent state initialization was also explored in the context of PINNs, with the physics-informed dynamics representation learner (PiDo)~\cite{wang_advancing_2025}, combining learnable initial latent codes with fully unsupervised physics-informed training. Additionally, in the same context, the Meta Auto-Decoder (MAD)~\cite{ye_meta-auto-decoder_2024} introduced the idea of improving auto-decoders with meta-learning, adopting a two-step training procedure where an initial pre-training phase is followed by a fine-tuning adaptation stage. Unlike these works, our approach adopts a data-driven supervised training that does not require explicit knowledge of the underlying PDE and can be seamlessly applied to experimental observations. Additionally, we explore a different meta-learning framework than the one in~\cite{ye_meta-auto-decoder_2024}. Indeed, our approach does not require multiple training stages and ensures fast adaptation to unseen tasks.

The contributions of this work are threefold. 
First, we formalize latent-state initialization from variable initial conditions as an encoder-free adaptation problem within the LDNet framework, thereby preserving its independence from spatial resolution and grid topology. 
Second, we introduce and systematically compare two encoder-free strategies, an auto-decoding formulation (AD-IC-LDNet) and a meta-learning formulation (Meta-IC-LDNet), across four benchmarks spanning advection-diffusion-reaction, fluid dynamics, and nonlinear solid mechanics. 
Third, in the test cases where both strategies are compared, we show that meta-learning accelerates latent-state inference by more than one order of magnitude, yields a better-conditioned adaptation loss, and leads to a latent space whose topology reflects that of the physical state space. 
Indeed, in all test cases, the organization of the meta-learned latent space reflects physically meaningful features of the underlying dynamics.

%% file: Sections/methods.tex
The Initial-Condition Latent Dynamics Network (IC-LDNet) overcomes one of the main limitations of the original LDNet architecture~\cite{regazzoni_learning_2024}
by accommodating varying initial conditions into the learning framework.
In this section, we first review the standard LDNet and its training procedure.
Subsequently, we introduce various strategies to inform the model about the initial state,
detailing the corresponding training paradigms. Finally, we introduce two latent-space regularization strategies.

\subsection{Problem}
\label{subsec:the-problem}
Given a spatial domain $\Omega\subset\mathbb{R}^d$ and a time
interval $[0,T]$, consider a dynamical system with input $\mathbf{u}:[0,T]\to U \subseteq \mathbb{R}^{d_u}$ and
output
$\mathbf{y}:\Omega\times[0,T]\to Y \subseteq \mathbb{R}^{d_y}$.
The goal is to learn the mapping between the input function $\mathbf{u}$ and the
output $\mathbf{y}$, provided that it is well-posed and causal
(i.e., the output at any time $t$ depends solely on the input history up to $t$).

\subsection{LDNet: architecture}
\label{subsubsec:the-ldnet-architecture}
The LDNet architecture consists of two coupled neural networks, the dynamics
network $\mathcal{NN}_{\mathrm{dyn}}$,
with parameters $\mathbf{w}_{\mathrm{dyn}}$, and the decoder network $\mathcal{NN}_{\mathrm{rec}}$,
with parameters $\mathbf{w}_{\mathrm{rec}}$.
Specifically, the LDNet evolves a latent state through a Neural Ordinary Differential Equation (Neural ODE)~\cite{chen_neural_2018}, and then
decodes it into the output space for each point in space and time.
The continuous model is summarized by the following dynamical system:
\begin{equation}
    \label{eq:ldnet-dynamical-system}
    \left\{
    \begin{aligned}
        \dot{\mathbf{s}}(t) &= \mathcal{NN}_{\mathrm{dyn}}(\mathbf{s}(t), \mathbf{u}(t); \mathbf{w}_{\mathrm{dyn}}) &\quad \text{for} \;
        t\in(0,T] \\
        \mathbf{s}(0) &= \mathbf{0} \\
        \tilde{\mathbf{y}}(\mathbf{x},t) &= \mathcal{NN}_{\mathrm{rec}}(\mathbf{s}(t), \mathbf{u}(t), \mathbf{x}; \mathbf{w}_{\mathrm{rec}})
        &\quad \text{for} \; \mathbf{x}\in\Omega \; \text{and} \; t\in[0,T],
    \end{aligned}
    \right.
\end{equation}
where $\mathbf{s}(t)\in\mathbb{R}^{d_s}$ is the vector of latent states at time $t$. In this work, we integrate the Neural ODE in time using a forward Euler scheme. 

\subsection{LDNet: training}
\label{subsubsec:training}
The LDNet is trained in a supervised manner~\cite{goodfellow_deep_2016}, given a training dataset of input and
output pairs $\mathcal{S}_{\mathrm{train}} = \{(\mathbf{u}_i(\tau), \mathbf{y}_i(\boldsymbol{\xi}, \tau))\}_{i=1}^{N_{\mathrm{tr}}}$, where
$\tau \in \mathcal{T}^i$ and $\boldsymbol{\xi} \in \mathcal{P}^i_{\tau}$, with $\mathcal{T}^i$ and $\mathcal{P}^i_{\tau}$
being the sets of time and space points, respectively, for the $i$-th training sample. These spatial points are not restricted to a specific grid or mesh, but can vary from sample to sample, and even from time point to time point.
The loss function is defined as:
\begin{equation}
    \label{eq:ldnet-loss-function}
    \mathcal{L}(\mathbf{w}_{\mathrm{dyn}}, \mathbf{w}_{\mathrm{rec}}) = \avsum_{i\in\mathcal{S}_{\mathrm{train}}} \avsum_{\tau\in
    \mathcal{T}^i}
    \avsum_{\boldsymbol{\xi}\in\mathcal{P}^i_{\tau}} \mathcal{E}
    \Big( \tilde{\mathbf{y}}_i(\boldsymbol{\xi}, \tau; \mathbf{w}_{\mathrm{dyn}}, \mathbf{w}_{\mathrm{rec}}), \mathbf{y}_i(
    \boldsymbol{\xi}, \tau) \Big) + \alpha_{\mathrm{dyn}} \mathcal{R}(\mathbf{w}_{\mathrm{dyn}}) + \alpha_{\mathrm{rec}} \mathcal{R}(\mathbf{w}_{\mathrm{rec}}),
\end{equation}
where the symbol $\avsum$ denotes the average operator, $\mathcal{E}$ is a suitable error measure (e.g., the squared
error), $\mathcal{R}$ is a regularization
term for the weights (e.g., $L_2$-regularization~\cite{goodfellow_deep_2016}), with weighting coefficients $\alpha_{\mathrm{dyn}}$ and $\alpha_{\mathrm{rec}}$.
An optimizer is then used to minimize the loss function with respect to the network parameters:
\begin{equation}
    \label{eq:ldnet-optimization-problem}
    (\mathbf{w}_{\mathrm{dyn}}^*, \mathbf{w}_{\mathrm{rec}}^*) = \argmin_{\mathbf{w}_{\mathrm{dyn}}, \mathbf{w}_{\mathrm{rec}}}
    \mathcal{L}(\mathbf{w}_{\mathrm{dyn}}, \mathbf{w}_{\mathrm{rec}}).
\end{equation}
Note that both $\mathcal{NN}_{\mathrm{dyn}}$ and $\mathcal{NN}_{\mathrm{rec}}$ are trained simultaneously, in an end-to-end fashion.
During inference, the trained LDNet can produce the output $\tilde{\mathbf{y}}(\mathbf{x}, t)$ (e.g., the approximate PDE solution) at any point in
space and time, given an input function $\mathbf{u}(t)$. Specifically, starting from $\mathbf{s}(0)=\mathbf{0}$, the latent state
is evolved by integrating in time the Neural ODE, and then decoded to the output space.

\subsection{Initial-Condition Latent Dynamics Network}
\label{subsec:initial-condition-latent-dynamics-network}
By leveraging the previously described LDNet architecture, it is possible to learn a surrogate reduced-order model for
time-dependent PDEs. 
However, as evident from Eq.~\eqref{eq:ldnet-dynamical-system},
the standard LDNet assumes that all trajectories start from the same initial condition, since the initial
latent state $\mathbf{s}(0)$ is always an arbitrary constant (e.g., a zero vector).
Note that the choice of a zero latent vector does not imply that the physical initial condition must be zero, since the decoder can learn to associate the fixed latent state with an arbitrary physical state. 
To overcome this limitation, we propose the Initial-Condition Latent Dynamics Network (IC-LDNet), embedding a set of initial observations into $\mathbf{s}(0)$.
The dynamical system is thus modified as follows:
\begin{equation}
    \label{eq:ic-ldnet-generic}
    \left\{
    \begin{aligned}
        \dot{\mathbf{s}}(t) &= \mathcal{NN}_{\mathrm{dyn}}(\mathbf{s}(t), \mathbf{u}(t); \mathbf{w}_{\mathrm{dyn}}) &\quad \text{for} \;
        t\in(0,T] \\
        \mathbf{s}(0) &= \mathbf{s}_0 = \Psi(\{ \mathbf{y}(\boldsymbol{\xi},\tau) \}_{\boldsymbol{\xi} \in \mathcal{P}_{\tau},\, \tau \in \mathcal{T}_{\mathrm{obs}}}; \mathbf{w}_{\mathrm{IC}})\\
        \tilde{\mathbf{y}}(\mathbf{x},t) &= \mathcal{NN}_{\mathrm{rec}}(\mathbf{s}(t), \mathbf{u}(t), \mathbf{x}; \mathbf{w}_{
                \mathrm{rec}}) &\quad \text{for} \; \mathbf{x}\in\Omega \; \text{and} \; t\in[0,T],
    \end{aligned}
    \right.
\end{equation}
where $\mathcal{T}_{\mathrm{obs}}$ denotes the set of initial observation times. The function $\Psi$, parametrized by $\mathbf{w}_{\mathrm{IC}}$, maps the system outputs available at these observation time steps to the
initial latent state $\mathbf{s}_0$. 
This generalized notation accommodates systems where only the snapshot at $t=0$ is considered, while also allowing for a short sequence of initial time steps to encode higher-order or non-Markovian dynamics when necessary. 
In the following
subsections, different strategies to define $\Psi$ are presented.

\subsubsection{IC-LDNet with encoder}
\label{subsubsec:ic-ldnet-with-encoder}
A straightforward approach to defining $\Psi$ is to use an encoder network, which maps the initial condition to the initial
latent state:
\begin{equation}
    \label{eq:ic-ldnet-encoder-initial-latent-state}
    \mathbf{s}_0 = \mathcal{NN}_{\mathrm{enc}}(\{ \mathbf{y}(\boldsymbol{\xi},\tau) \}_{\boldsymbol{\xi} \in \mathcal{P}_{\tau},\, \tau \in \mathcal{T}_{\mathrm{obs}}}; \mathbf{w}_{\mathrm{IC}}),
\end{equation}
where $\mathcal{NN}_{\mathrm{enc}}$ is a generic encoder network with parameters $\mathbf{w}_{\mathrm{IC}}$.
The encoder can be jointly trained with the dynamics and decoder networks in an end-to-end
fashion. At inference time, the initial condition is then passed through the encoder to obtain $\mathbf{s}_0$, and
then the latent state is evolved and decoded as in the original implementation.
However, standard encoder architectures are tied to a specific data format (e.g., convolutional networks~\cite{li_survey_2022} for regular grids, graph neural networks~\cite{hamilton_graph_2020} for unstructured meshes), which would compromise the discretization invariance of the LDNet.
Although discretization-agnostic encoders exist (e.g., attention-based architectures~\cite{serrano_aroma_2024}), the introduction of an additional network, with respect to the original LDNet, substantially increases the number of parameters and thus the computational and memory cost of training and inference.
Hence, in this work, the IC-LDNet with encoder is not implemented, in favor of
encoder-free strategies.

\subsubsection{IC-LDNet with auto-decoder}
\label{subsubsec:ic-ldnet-with-auto-decoder}
An alternative strategy to define $\Psi$ is to adopt a decoder-only or \emph{auto-decoder} approach~\cite{tan_reducing_1995,bojanowski_optimizing_2018, park_deepsdf_2019,wang_advancing_2025}. In this case, $\Psi$ associates each initial condition in the training set with a specific trainable vector. Specifically, given the training
set $\mathcal{S}_{\mathrm{train}} = \{(\mathbf{u}_i(\tau), \mathbf{y}_i(\boldsymbol{\xi}, \tau))\}_{i=1}^{N_{\mathrm{tr}}}$, the
initial latent state for the $i$-th training sample is defined as:
\begin{equation}
    \label{eq:ic-ldnet-auto-decoder-initial-latent-state}
    \mathbf{s}_{0}^{(i)} = \mathbf{e}_i^T\mathbf{W}_{\mathrm{IC}}^{\mathrm{train}} \quad \text{with} \quad
    \mathbf{W}_{\mathrm{IC}}^{\mathrm{train}} \in \mathbb{R}^{N_{\mathrm{tr}} \times d_s},
\end{equation}
where $\mathbf{W}_{\mathrm{IC}}^{\mathrm{train}}$ stores all the initial latent codes for the training set, and $\mathbf{e}_i$ is the $i$-th element of the canonical basis of $\mathbb{R}^{N_{\mathrm{tr}}}$. In particular, $\mathbf{W}_{\mathrm{IC}}^{\mathrm{train}}$ is initialized by sampling from a Gaussian distribution $\mathcal{N}(0, 10^{-4})$ 
and optimized
jointly with the network parameters.
Hence, the loss function becomes:
\begin{equation}
    \label{eq:ic-ldnet-auto-decoder-loss-fn}
    \begin{aligned}
        \mathcal{L}(\mathbf{w}_{\mathrm{dyn}}, \mathbf{w}_{\mathrm{rec}}, \mathbf{W}_{\mathrm{IC}}^{\mathrm{train}}) &=
        \avsum_{i\in\mathcal{S}_{\mathrm{train}}} \avsum_{\tau\in\mathcal{T}^i} \avsum_{\boldsymbol{\xi}\in\mathcal{P}^i_{\tau}}
        \mathcal{E}
        \Big(
        \tilde{\mathbf{y}}_i(\boldsymbol{\xi}, \tau; \mathbf{w}_{\mathrm{dyn}}, \mathbf{w}_{\mathrm{rec}}, \mathbf{W}_{\mathrm{IC}}^{\mathrm{train}}),
        \mathbf{y}_i( \boldsymbol{\xi}, \tau)
        \Big) \\
        & \quad +
        \alpha_{\mathrm{dyn}} \mathcal{R}(\mathbf{w}_{\mathrm{dyn}}) +
        \alpha_{\mathrm{rec}} \mathcal{R}(\mathbf{w}_{\mathrm{rec}}) +
        \alpha_{\mathrm{IC}} \mathcal{R}(\mathbf{W}_{\mathrm{IC}}^{\mathrm{train}}),
    \end{aligned}
\end{equation}
where $\mathcal{R}(\mathbf{W}_{\mathrm{IC}}^{\mathrm{train}})$ is a regularization term with weight $\alpha_{\mathrm{IC}}$. The optimization problem is thus:
\begin{equation}
    \label{eq:ic-ldnet-auto-decoder-optimization}
        (\mathbf{w}_{\mathrm{dyn}}^*, \mathbf{w}_{\mathrm{rec}}^*, \mathbf{W}_{\mathrm{IC}}^{\mathrm{train}\;*}) = \argmin_{\mathbf{w}_{\mathrm{dyn}},
    \mathbf{w}_{\mathrm{rec}}, \mathbf{W}_{\mathrm{IC}}^{\mathrm{train}}} \mathcal{L}(
    \mathbf{w}_{\mathrm{dyn}}, \mathbf{w}_{\mathrm{rec}}, \mathbf{W}_{\mathrm{IC}}^{\mathrm{train}}).
\end{equation}
This approach has the advantage of being encoder-free, thus avoiding the previously mentioned drawbacks.
However, at inference time, it is necessary to fit a new latent code for the initial condition of each new example in the evaluation set $\mathcal{S}_{\mathrm{eval}} = \{(\mathbf{u}_i(\tau), \mathbf{y}_i(\boldsymbol{\xi}, \tau))\}_{i=1}^{N_{\mathrm{ev}}}$.
Specifically, we initialize the collection of codes for the new initial conditions (i.e., $\mathbf{W}_{\mathrm{IC}}^{\mathrm{eval}} \in \mathbb{R}^{N_{\mathrm{ev}} \times d_s}$) by sampling from the same Gaussian distribution as before. 
Then, after freezing the trained network parameters $(\mathbf{w}_{\mathrm{dyn}}^*, \mathbf{w}_{\mathrm{rec}}^*)$, we minimize the following loss function with respect to the new latent codes:
\begin{equation}
    \label{eq:ic-ldnet-auto-decoder-inference-loss-fn}
        \mathcal{L}(\mathbf{W}_{\mathrm{IC}}^{\mathrm{eval}}) =
        \avsum_{i\in\mathcal{S}_{\mathrm{eval}}} \avsum_{\tau\in\mathcal{T}_{\mathrm{eval}}^i} \avsum_{\boldsymbol{\xi}\in\mathcal{P}
    ^i_{\tau}}
        \mathcal{E}
        \Big(
        \tilde{\mathbf{y}}_i(\boldsymbol{\xi}, \tau; \mathbf{W}_{\mathrm{IC}}^{\mathrm{eval}}),
        \mathbf{y}_i( \boldsymbol{\xi}, \tau)
        \Big) +
        \alpha_{\mathrm{IC}} \mathcal{R}(\mathbf{W}_{\mathrm{IC}}^{\mathrm{eval}}).
\end{equation}
Notably, $\mathcal{T}_{\mathrm{eval}}^i \subseteq [0, T_{\mathrm{eval}}]$, with $T_{\mathrm{eval}} \ll T$, since the initial latent state should be inferred from only a short initial portion of the trajectory. In the simplest case of $T_{\mathrm{eval}}=0$, only the initial condition is used. For non-Markovian systems, however, a short time window with $T_{\mathrm{eval}}>0$ may be required to provide sufficient information about the system state.
Once optimized, these initial latent codes are evolved through the Neural ODE and decoded to predict the entire system trajectory.
In the rest of this work, this approach is referred to as the Auto-Decoder IC-LDNet, or AD-IC-LDNet.

\begin{algorithm}[b!]
	\caption{Meta-IC-LDNet: training procedure}
    \label{alg:cavia-train}
	\begin{algorithmic}[1]
		\Require Training dataset $\mathcal{S}_{\mathrm{train}} = \{(\mathbf{u}_i(\tau), \mathbf{y}_i(\boldsymbol{\xi}, \tau))\}_{i=1}^{N_{\mathrm{tr}}}$
		\Require Initial meta-parameters $\boldsymbol{\theta} = \{\mathbf{w}_{\mathrm{dyn}}, \mathbf{w}_{\mathrm{rec}}\}$, randomly initialized
        \Require Learning rates $\alpha$ and $\beta$, and number of inner-loop iterations $K$
        \Require Inner loss
        $\mathcal{L}_{i}^{\mathrm{I}}\Big(\boldsymbol{\theta}, \mathbf{s}_{0}^{(i)}\Big) =
        \avsum_{\tau\in\mathcal{T}^i_{\mathrm{inner}}} \avsum_{\boldsymbol{\xi}\in\mathcal{P}^i_{\tau}}
        \mathcal{E}
        \Big(
        \tilde{\mathbf{y}}_i(\boldsymbol{\xi}, \tau; \boldsymbol{\theta}, \mathbf{s}_{0}^{(i)}),
        \mathbf{y}_i( \boldsymbol{\xi}, \tau)
        \Big)$ with $i\in [1, N_{\mathrm{tr}}]$
        \Require Outer loss
        $\mathcal{L}_{i}^{\mathrm{O}}\Big(\boldsymbol{\theta}, \mathbf{s}_{0}^{(i)}\Big) =
        \avsum_{\tau\in\mathcal{T}^i_{\mathrm{outer}}} \avsum_{\boldsymbol{\xi}\in\mathcal{P}^i_{\tau}}
        \mathcal{E}
        \Big(
        \tilde{\mathbf{y}}_i(\boldsymbol{\xi}, \tau; \boldsymbol{\theta}, \mathbf{s}_{0}^{(i)}),
        \mathbf{y}_i( \boldsymbol{\xi}, \tau)
        \Big)$ with $i\in [1, N_{\mathrm{tr}}]$
		\While{not converged}
            \For{$i \in \{1, \ldots, N_{\mathrm{tr}}\}$}
                \State Initialize task-specific parameters: $\mathbf{s}_{0}^{(i, 0)} \gets \mathbf{0}$ \Comment{Reset codes}
                \For{$k \in \{1, \ldots, K\}$}
                    \State $\mathbf{s}_{0}^{(i, k)} \gets \mathbf{s}_{0}^{(i, k-1)} - \alpha \nabla_{\mathbf{s}_{0}^{(i, k-1)}} \mathcal{L}_{i}^{\mathrm{I}}\Big(\boldsymbol{\theta}, \mathbf{s}_{0}^{(i, k-1)}\Big)$ \Comment{Find optimal $\mathbf{s}_{0}$}

                \EndFor
            \EndFor
            \State $\boldsymbol{\theta} \gets \boldsymbol{\theta} - \beta \nabla_{\boldsymbol{\theta}}
                    \avsum_{i=1}^{N_{\mathrm{tr}}} \mathcal{L}_{i}^{\mathrm{O}}\Big(\boldsymbol{\theta}, \mathbf{s}_{0}^{(i, K)}(\boldsymbol{\theta})\Big)$ \Comment{Update meta-parameters}
		\EndWhile
	\end{algorithmic}
\end{algorithm}
\subsubsection{IC-LDNet with meta-learning}
\label{subsubsec:ic-ldnet-with-meta-learning}
A second encoder-free strategy relies on meta-learning~\cite{hospedales_meta-learning_2022,ye_meta-auto-decoder_2024}, 
specifically on the CAVIA framework~\cite{zintgraf_fast_2019}, following the idea presented for
CORAL~\cite{serrano_operator_2023}. While CORAL applies CAVIA-style meta-learning to a two-stage encode-process-decode pipeline, we integrate it into the end-to-end LDNet training scheme, where latent dynamics and reconstruction are learned jointly rather than sequentially.
The overall structure is similar to the IC-LDNet
with auto-decoder presented in Section~\ref{subsubsec:ic-ldnet-with-auto-decoder}, but it differs in the training procedure.
Specifically, $\mathbf{s}_0^{(i)}$ represents a set of context parameters specific to the $i$-th training
example,
while $\boldsymbol{\theta} = \{\mathbf{w}_{\mathrm{dyn}}, \mathbf{w}_{\mathrm{rec}}\}$ are the meta-parameters, shared across all tasks.
In the context of the
IC-LDNet, the training procedure is reported in Algorithm~\ref{alg:cavia-train}. Note that the gradient with respect to the
meta-parameters is computed by back-propagating through the inner loop, since $\mathbf{s}_{0}^{(i, K)}$ depends
on $\boldsymbol{\theta}$. Moreover, meta-parameter update can be performed using more sophisticated approaches than gradient descent, such as Adam~\cite{kingma_adam_2017}.

In the original CAVIA implementation, the training batch is divided into two
parts, one used in the inner loop and one in the final gradient step~\cite{zintgraf_fast_2019}. In contrast, the CORAL
architecture uses the entire batch for both phases~\cite{serrano_operator_2023}.
In this work, in the inner loop, the loss function is computed only on a short
time interval $[0, T_{\mathrm{inner}}^{\mathrm{train}}]$, with $T_{\mathrm{inner}}^{\mathrm{train}} \ll T$, and at discrete times $\tau \in \mathcal{T}_{\mathrm{inner}}^i \subseteq [0, T_{\mathrm{inner}}^{\mathrm{train}}]$.
After the inner loop, the meta-parameters are updated using the loss computed
on the entire trajectory (i.e., $\tau \in \mathcal{T}_{\mathrm{outer}}^i \subseteq [0, T]$).
Alternatively, the outer loss function can be computed only on the remainder of the trajectory (i.e., $\tau \in \mathcal{T}_{\mathrm{outer}}^i \subseteq (T_{\mathrm{inner}}^{\mathrm{train}}, T]$). For each test case in the results section, the specific choice of the outer-loss time domain will be specified.
At inference time, as summarized in Algorithm~\ref{alg:cavia-eval}, the
meta-parameters $\boldsymbol{\theta}$ are fixed, and only the context parameters $\mathbf{s}_{0}^{(i)}$ are updated.
Similarly to the training procedure, the inner loss is computed only on a short time interval
$[0, T_{\mathrm{inner}}^{\mathrm{eval}}]$ to fit the
context
parameters to the
initial condition. For simplicity, the same notation as in the training procedure is used, with
$\mathcal{T}_{\mathrm{inner}}^i \subseteq [0, T_{\mathrm{inner}}^{\mathrm{eval}}]$.
After the inner loop, the optimized context parameters are used to predict the entire trajectory.
Importantly, in the meta-learning approach, the procedure used to infer unseen initial latent states is already embedded in the training process, unlike in the AD-IC-LDNet. This provides the basis for a faster and more robust adaptation to unseen initial conditions. In this work, this approach is called Meta-IC-LDNet.

\begin{algorithm}[t!]
	\caption{Meta-IC-LDNet: inference procedure}
    \label{alg:cavia-eval}
	\begin{algorithmic}[1]
		\Require Evaluation dataset $\mathcal{S}_{\mathrm{eval}} = \{(\mathbf{u}_i(\tau), \mathbf{y}_i(\boldsymbol{\xi}, \tau))\}_{i=1}^{N_{\mathrm{ev}}}$
		\Require Trained meta-parameters $\boldsymbol{\theta} = \{\mathbf{w}_{\mathrm{dyn}}, \mathbf{w}_{\mathrm{rec}}\}$
        \Require Learning rate $\alpha$, and number of inner-loop iterations $K$
        \Require Inner loss
        $\mathcal{L}_{i}^{\mathrm{I}}\Big(\boldsymbol{\theta}, \mathbf{s}_{0}^{(i)}\Big) =
        \avsum_{\tau\in\mathcal{T}_{\mathrm{inner}}^i} \avsum_{\boldsymbol{\xi}\in\mathcal{P}^i_{\tau}}
        \mathcal{E}
        \Big(
        \tilde{\mathbf{y}}_i(\boldsymbol{\xi}, \tau; \boldsymbol{\theta}, \mathbf{s}_{0}^{(i)}),
        \mathbf{y}_i( \boldsymbol{\xi}, \tau)
        \Big)$ with $i\in [1, N_{\mathrm{ev}}]$
        \For{$i \in \{1, \ldots, N_{\mathrm{ev}}\}$}
            \State Initialize task-specific parameters: $\mathbf{s}_{0}^{(i, 0)} \gets \mathbf{0}$ \Comment{Initialize codes}
            \For{$k \in \{1, \ldots, K\}$}
                \State
        $\mathbf{s}_{0}^{(i, k)} \gets \mathbf{s}_{0}^{(i, k-1)} - \alpha \nabla_{\mathbf{s}_{0}^{(i, k-1)}} \mathcal{L}_{i}^{\mathrm{I}}\Big(\boldsymbol{\theta}, \mathbf{s}_{0}^{(i, k-1)}\Big)$ \Comment{Find optimal $\mathbf{s}_{0}$}
            \EndFor
            \State Compute $\tilde{\mathbf{y}}_i (\mathbf{x}, t; \boldsymbol{\theta}, \mathbf{s}_{0}^{(i, K)})$ with $\mathbf{x}\in\Omega$ and $t\in[0, T]$ \Comment{Predict spatio-temporal dynamics}
        \EndFor
	\end{algorithmic}
\end{algorithm}

\subsubsection{Regularization strategies for training stability and latent-space uniqueness}
\label{subsubsec:regularization-stability}
Depending on the specific test case, the IC-LDNet may suffer from training
instability and may produce non-unique latent representations. Specifically, the former issue is mainly observed when training the
Meta-IC-LDNet, possibly due to the reset of the context parameters at each iteration. Instead, the latter is particularly problematic for the AD-IC-LDNet, due to the random initialization of the trainable initial codes. We therefore propose two regularization strategies. For each of the four test cases in Section~\ref{sec:results}, we discuss whether regularization is needed and, if so, which strategy is adopted.

\paragraph{Curriculum learning}
\label{par:curriculum_learning}
The first strategy is to introduce a \emph{continuation} or \emph{curriculum-learning} approach~\cite{bengio_curriculum_2009},
where the loss function (the outer one for the meta-learning approach) is computed on a progressively longer time horizon as training proceeds. Indeed, the
random initialization of the dynamics network, combined with long time horizons, may initially produce diverging
trajectories that hinder training. This strategy requires a careful choice of hyperparameters,
such as the initial time horizon and the rate of increase. Moreover, it requires longer training, for the
model gradually learns to predict longer trajectories.

\paragraph{Latent-state penalty}
\label{par:latent_state_penalty}
The second strategy is to actively prevent the latent trajectory from diverging excessively from the initial state. To achieve this,
a straightforward approach is to add an $L_2$ penalty on the final latent state, with a suitable
weighting coefficient. For the Meta-IC-LDNet, a more sophisticated approach is to penalize the
distance (e.g., Mean Squared Error, MSE) between a subset of the trajectory states and the latent codes that would be obtained by
inner-loop optimization at the same time instants. The advantage of the latter approach is that the penalty term
converges to zero in the limit of perfect optimization.
However, it requires additional inner-loop optimization steps, increasing training time and cost.

%% file: Sections/results.tex
To demonstrate the performance of the IC-LDNet and compare the auto-decoder and meta-learning frameworks, the proposed methods are evaluated on four benchmark problems, as summarized in Table~\ref{tab:nrmse-results}. Due to the observed advantages of the Meta-IC-LDNet over the AD-IC-LDNet, we tackle the last two test cases with just the former approach.
These tests are designed to thoroughly assess the model predictive accuracy and generalization capabilities across diverse physical domains. 
Unless otherwise specified, training is performed on a single custom NVIDIA A100 GPU (64 GB VRAM) hosted on CINECA's LEONARDO HPC cluster~\cite{turisini_leonardo_2024},
and the training budgets and inference times are summarized in Table~\ref{tab:computational-costs}.

For each test case, we performed a preliminary hyperparameter-tuning campaign, where several combinations of architectural and optimization parameters were tested to find those that correspond to the best generalization on the validation set. Additionally, the robustness of the hyperparameter combinations to the choice of the specific random seeds was verified during tuning. Furthermore, the test set was always set aside for just the final inference. In Appendix~\ref{sec:appendix-hyper}, the final hyperparameter configurations are reported for each test case and IC-LDNet variant.

\begin{table}[htbp]
    \centering
    \small
    \renewcommand{\arraystretch}{1}
    \begin{tabular}{llccc}
        \toprule
        \textbf{Test case} & \textbf{Architecture} & \multicolumn{3}{c}{\textbf{NRMSE (Test set)}} \\
        \midrule
        
        % --- ADR ---
        \multirow{2}{*}{ADR} & AD-IC-LDNet & \multicolumn{3}{c}{$3.5 \times 10^{-3}$} \\
         & Meta-IC-LDNet & \multicolumn{3}{c}{$2.9 \times 10^{-3}$} \\
        \midrule
        
        % --- Static Cylinder ---
        & & $p$ & $u$ & $v$ \\
        \cmidrule{3-5}
        \multirow{2}{*}{Static cylinder} & AD-IC-LDNet & $4.1 \times 10^{-3}$ & $4.7 \times 10^{-3}$ & $7.1 \times 10^{-3}$ \\
         & Meta-IC-LDNet & $2.6 \times 10^{-3}$ & $2.7 \times 10^{-3}$ & $4.1 \times 10^{-3}$ \\
        \midrule
        
        % --- Rotating Cylinder ---
        & & $p$ & $u$ & $v$ \\
        \cmidrule{3-5}
        Rotating cylinder & Meta-IC-LDNet & $7.5 \times 10^{-3}$ & $1.2 \times 10^{-2}$ & $1.3 \times 10^{-2}$ \\
        \midrule
        
        % --- Nonlinear Beam ---
        & & $d_x$ & $d_y$ & $d_z$ \\
        \cmidrule{3-5}
        Nonlinear beam & Meta-IC-LDNet & $4.8 \times 10^{-2}$ & $4.7 \times 10^{-2}$ & $4.5 \times 10^{-2}$ \\
        
        \bottomrule
    \end{tabular}
    \caption{Comparison of per-component test-set range-Normalized Root Mean Square Error (NRMSE) across the four test cases.}
    \label{tab:nrmse-results}
\end{table}

\begin{table}[htbp]
\centering
\renewcommand{\arraystretch}{1.}
\small
\begin{threeparttable}
\begin{tabular}{llccc}
\toprule
\textbf{Test case} & \textbf{Architecture} & \makecell{\textbf{Hardware} \\ \textbf{(Training budget)}} & \makecell{\textbf{Inference} \\ \textbf{mini-batch size}} & \makecell{\textbf{Inference} \\ \textbf{time}\tnote{a}} \\
\midrule
\multirow{2}{*}{ADR} & AD-IC-LDNet & HPC cluster ($<$24 h)\tnote{b} & 100 & 1.14 s/ex. \\
 & Meta-IC-LDNet & HPC cluster (24 h) & 100 & 0.04 s/ex. \\
\addlinespace
\multirow{2}{*}{Static cylinder} & AD-IC-LDNet & HPC cluster (24 h) & 6 & 21.11 s/ex. \\
 & Meta-IC-LDNet & HPC cluster (24 h) & 6 & 0.23 s/ex. \\
\addlinespace
Rotating cylinder & Meta-IC-LDNet & HPC cluster (24 h)\tnote{c} & 2 & 0.87 s/ex. \\
\addlinespace
Nonlinear beam & Meta-IC-LDNet & Workstation (Convergence)\tnote{d} & 8 & 0.15 s/ex. \\
\bottomrule
\end{tabular}
\begin{tablenotes}[flushleft]
\footnotesize
\item[] Unless otherwise noted, HPC training runs were executed until the end of the 24-hour queue.
\item[a] Measured in seconds per example (s/ex.) on the test set.
\item[b] The loss reached a plateau before the 24 h HPC queue limit and training was interrupted.
\item[c] Fine-tuning is then performed for an additional 24 hours.
\item[d] No fixed training budget was set: training was run until observed convergence of the losses (see Appendix~\ref{sec:appendix-train-val}).
\end{tablenotes}
\end{threeparttable}
\caption{Training hardware and inference speed for different test cases and architectures.}
\label{tab:computational-costs}
\end{table}

\subsection{Test case 1: 1D advection-diffusion-reaction}
\label{subsec:test-case-adr}
\input{Sections/result_subsec/adr}

\subsection{Test case 2A: 2D flow around a static circular cylinder}
\label{subsec:test-case-cyl}
\input{Sections/result_subsec/cylinder}

\subsection{Test case 2B: 2D flow around a rotating circular cylinder}
\label{subsec:test-case-randomrot}
\input{Sections/result_subsec/randomrot}

\subsection{Test case 3: nonlinear solid dynamics of a 3D cantilever beam}
\label{subsec:test-case-beam}
\input{Sections/result_subsec/beam}

%% file: Sections/result_subsec/adr.tex
The Advection-Diffusion-Reaction (ADR) equation is a second-order PDE
that
provides a fundamental benchmark for evaluating the performance of neural networks in approximating PDE solutions. Physically,
the solution can be interpreted as the concentration of a substance being transported and diffused in a medium,
while also undergoing
a reaction process~\cite{quarteroni_numerical_2014}.

\noindent Specifically, we consider the following
differential 1D problem:
\begin{equation}
    \label{eq:adr-equation}
    \begin{cases}
        \displaystyle \frac{\partial z(x, t)}{\partial t} - \mu_1 \frac{\partial^2 z(x, t)}{\partial x^2} - \mu_2
        \frac{\partial z(x, t)}{\partial x} + \mu_3 z(x, t) = f(x, t) &\quad x \in [-1, 1],\; t \in (0, T] \\
        z(x, 0) = z_0(x) &\quad x \in [-1, 1] \\
        \text{(Periodic BCs)}, &
    \end{cases}
\end{equation}
with constants $\mu_1, \mu_2, \mu_3$ defining the diffusion, advection, and reaction coefficients, respectively.
To create
the dataset,
the PDE coefficients (i.e., constant inputs $\mathbf{u}(t)$) are randomly sampled as follows:
\begin{equation}
    \label{eq:sampled-adr-params}
    \left\{
        \begin{aligned}
            \mu_1 &\sim \mathcal{U}(0, 0.05) \\
            \mu_2 &\sim \mathcal{U}(-0.1, 0.1) \\
            \mu_3 &\sim \mathcal{U}(0, 0.01),
        \end{aligned}
    \right.
\end{equation}
with $\mathcal{U}(a, b)$ denoting the uniform distribution in the interval $[a, b]$.
The initial conditions, on the other hand, are defined as sine waves with randomly sampled amplitude and phase shift:
\begin{equation}
    \label{eq:sampled-adr-initial-conds}
    z_0(x) = A \sin(\pi x + \varphi),
\end{equation}
where the parameters are sampled as:
\begin{equation}
    \label{eq:sampled-adr-initial-conds-params}
    \left\{
        \begin{aligned}
            A &\sim \mathcal{U}(0.5, 1.5) \\
            \varphi &\sim \mathcal{U}(-\pi, \pi).
        \end{aligned}
    \right.
\end{equation}
Given $f(x, t)=0$ and the initial condition defined in~\eqref{eq:sampled-adr-initial-conds},
the PDE in~\eqref{eq:adr-equation} admits the following analytical solution:
\begin{equation}
    \label{eq:adr-true-solution}
    z(x, t) = A \exp\left(-(\mu_1 \pi^2+\mu_3)t\right) \sin(\pi (x + \mu_2 t) + \varphi).
\end{equation}
Equation~\eqref{eq:adr-true-solution} shows that the evolution of the problem can be fully described by two states (i.e., amplitude and phase).
Although the analytical expression of the solution is available,
we generate the dataset using the same FFT-based solver adopted in the original LDNet implementation~\cite{regazzoni_learning_2024}.
Specifically, the dataset comprises 1000 samples; for each sample, the solution is evaluated at 101 equally spaced spatial
points and 100 time steps. The dataset is subsequently randomly split into training, validation, and test sets, with 800, 100,
and 100 samples, respectively.

\begin{figure}[ht!]
    \centering
    \begin{subfigure}{\textwidth}
        \centering
        \includegraphics[scale=1.]{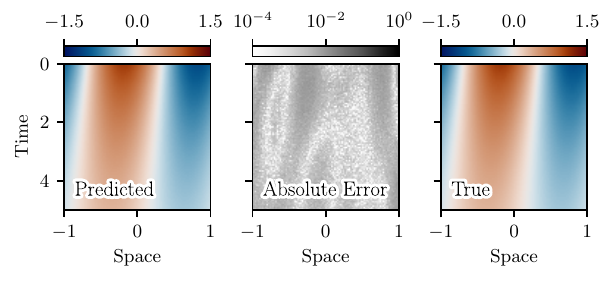}
        \caption{Best prediction
            (NRMSE $1.6 \times 10^{-3}$).
        }
        \label{subfig:autodec_best_pred_adr}
    \end{subfigure}
    \begin{subfigure}{\textwidth}
        \centering
        \includegraphics[scale=1.]{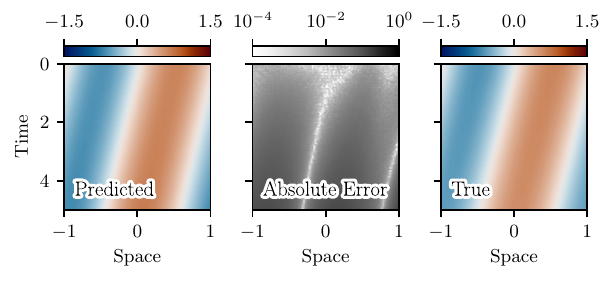}
        \caption{Worst prediction
            (NRMSE $3.5 \times 10^{-2}$).
        }
        \label{subfig:autodec_worst_pred_adr}
    \end{subfigure}
    \caption{ADR test case: inference on the test set (AD-IC-LDNet).
    Predicted solution, absolute error, and true solution in the space-time domain for the test-set samples with the lowest (a) and highest (b) NRMSE.}
    \label{fig:autodec-inference-adr}
\end{figure}

\begin{figure}[b!]
    \centering
    \includegraphics[scale=1.]{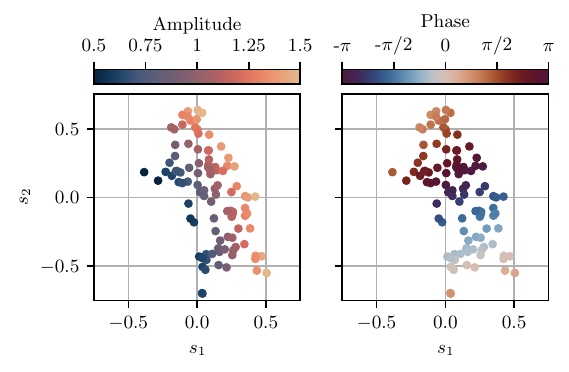}
    \caption{ADR test case: initial latent states (AD-IC-LDNet).
    Test-set initial codes, colored by the amplitude (left) and the phase (right) of the corresponding initial conditions.}
    \label{fig:autodec-initial-latent-adr}
\end{figure}

\subsubsection{AD-IC-LDNet on the ADR test case}
\label{subsubsec:adr-autodec}

We first train the AD-IC-LDNet surrogate defined in Section~\ref{subsubsec:ic-ldnet-with-auto-decoder}. In this case, the architecture of the model
and the hyperparameters are similar to those used in the original LDNet
paper~\cite{regazzoni_learning_2024} and are summarized in Table~\ref{tab:hyperparameters-comparison-adr} (see Appendix~\ref{sec:appendix-hyper}). Specifically, the latent
dimension is $d_s=2$, which corresponds to the intrinsic dimension of the solution manifold, as observed previously.
The trained model is evaluated on the test set by first optimizing the initial latent states of the unseen samples, followed by the evolution of the latent dynamics and the reconstruction to the physical space.
In this context, in Figures~\ref{subfig:autodec_best_pred_adr}
and~\ref{subfig:autodec_worst_pred_adr}, respectively, we present the best and worst test-set predictions in terms of range-Normalized Root Mean Square Error (NRMSE).
Interestingly, the predictive performance varies considerably between samples, with the worst sample characterized by an NRMSE of $3.5 \times 10^{-2}$, significantly higher than the $1.6 \times 10^{-3}$ of the best sample.
However, an average test-set NRMSE of $3.5 \times 10^{-3}$ demonstrates that the AD-IC-LDNet still accurately predicts the ADR dynamics for the majority of samples.

\begin{figure}[t!]
    \centering
    \includegraphics[scale=1.]{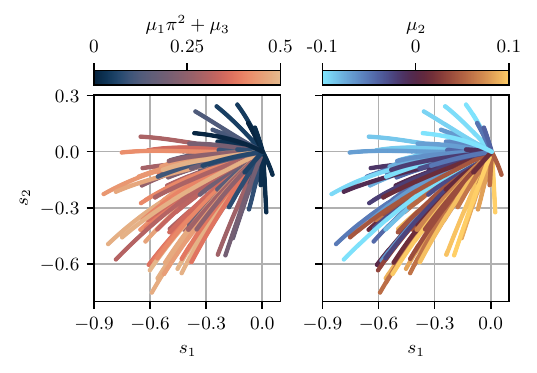}
    \caption{ADR test case: latent trajectories (AD-IC-LDNet). Test-set latent trajectories, translated to start from the origin, colored by $\mu_1\pi^2+\mu_3$, which combines diffusion and reaction (left), and by the advection coefficient $\mu_2$ (right).}
    \label{fig:autodec-latent-traj-adr}
\end{figure}

Additionally, we analyze how the learned latent space is organized.
As anticipated,
a latent dimension of two would have a natural correspondence with the amplitude and phase of the exact solution.
However, it is not guaranteed that the learned latent variables will directly correspond to these physical parameters, but rather to one of the infinitely many equivalent parametrizations.
To investigate this, in Figure~\ref{fig:autodec-initial-latent-adr}, the initial test-set states are
plotted in the 2D latent space,
colored according to the amplitude and phase of the corresponding initial conditions. Interestingly,
we observe that the two physical parameters vary along two virtually perpendicular directions. However,
the latent codes are not periodically arranged according to the phase, as an optimal encoding
reflecting the topology of the physical space would require.
Indeed, the learned representation is redundant,
as a single state in the physical space may admit two alternative representations in the latent space.
Finally, Figure~\ref{fig:autodec-latent-traj-adr} shows the link between the latent
trajectories, moved to a common starting point,
and the PDE coefficients.
Here, we observe that the combined effect of diffusion and reaction is dominant in examples that move in the amplitude direction,
while the advection coefficient influences the drift in the phase direction. This is in accordance with the physical interpretation of the analytical solution in Equation~\eqref{eq:adr-true-solution}.

\begin{figure}[ht!]
    \centering
    \begin{subfigure}{\textwidth}
        \centering
        \includegraphics[scale=1.]{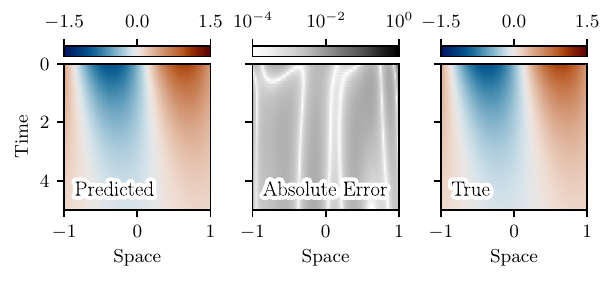}
        \caption{Best prediction (NRMSE $1.4 \times 10^{-3}$).}
        \label{subfig:cavia_best_pred_adr}
    \end{subfigure}
    \begin{subfigure}{\textwidth}
        \centering
        \includegraphics[scale=1.]{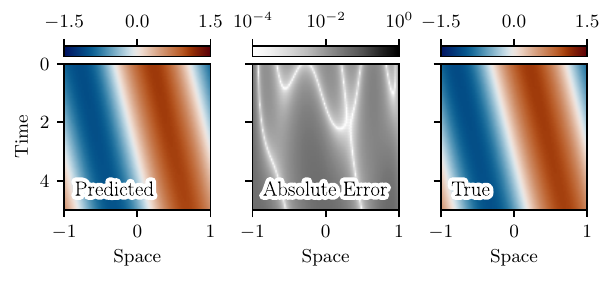}
        \caption{Worst prediction (NRMSE $8.0 \times 10^{-3}$).}
        \label{subfig:cavia_worst_pred_adr}
    \end{subfigure}
    \caption{ADR test case: inference on the test set (Meta-IC-LDNet).
    Predicted solution, absolute error, and true solution in the space-time domain for the test-set samples with the lowest (a) and highest (b) NRMSE.}
    \label{fig:cavia-inference-adr}
\end{figure}

\subsubsection{Meta-IC-LDNet on the ADR test case}
\label{subsubsec:adr-cavia}
As an alternative to the AD-IC-LDNet, we investigate the Meta-IC-LDNet described in Section~\ref{subsubsec:ic-ldnet-with-meta-learning}. To have a fair comparison, the number of parameters of $\mathcal{NN}_{\mathrm{dyn}}$ and $\mathcal{NN}_{\mathrm{rec}}$, as well as the latent dimension $d_s$,
are kept the same as in Section~\ref{subsubsec:adr-autodec} and in the original LDNet
paper~\cite{regazzoni_learning_2024}. Here, the main difference is the use of SiLU activations~\cite{elfwing_sigmoid-weighted_2018}, due to training issues with the original tanh. Overall, the full set of hyperparameters is summarized in Table~\ref{tab:hyperparameters-comparison-adr} in Appendix~\ref{sec:appendix-hyper}.
With respect to Algorithm~\ref{alg:cavia-train}, the inner loss $\mathcal{L}_{i}^{\mathrm{I}}$ is computed only on the initial condition (i.e., $T_{\mathrm{inner}}^{\mathrm{train}}=0$), while the outer loss $\mathcal{L}_{i}^{\mathrm{O}}$ considers the remainder of the trajectory (i.e., $\mathcal{T}_{\mathrm{outer}}^i = (0, T]$).

Following Algorithm~\ref{alg:cavia-eval}, the model is evaluated on the test set,
and Figure~\ref{fig:cavia-inference-adr} shows the best and worst predictions in terms of NRMSE.
By comparing Figures~\ref{subfig:autodec_worst_pred_adr} and~\ref{subfig:cavia_worst_pred_adr},
we can observe that the worst prediction of the Meta-IC-LDNet is considerably more accurate than that of the AD-IC-LDNet. Remarkably, the meta-learning approach leads to a more evenly distributed error across the test set and an overall better accuracy, with an average NRMSE of $2.9 \times 10^{-3}$.

\begin{figure}[ht!]
    \centering
    \begin{subfigure}{0.6\textwidth}
        \centering
        \includegraphics[scale=1.]{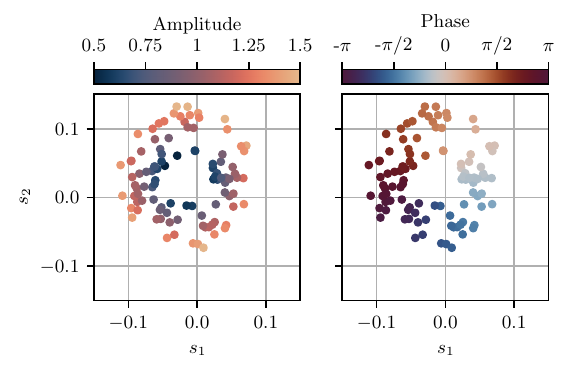}
        \caption{Initial latent states.}
        \label{subfig:cavia-initial-latent-adr}
    \end{subfigure}
    \begin{subfigure}{0.39\textwidth}
        \centering
        \includegraphics[scale=1.]{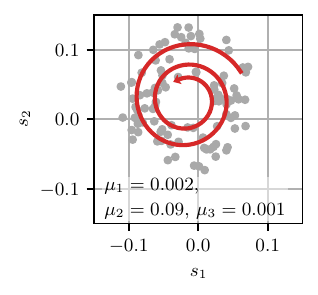}
        \caption{Advection-dominated trajectory.}
        \label{subfig:cavia-advection-adr}
    \end{subfigure}
    \caption{ADR test case: latent space (Meta-IC-LDNet).
        (a) Test-set initial codes, colored by the amplitude (left) and the phase (right) of the corresponding initial conditions.
        (b) Latent trajectory (red) of an advection-dominated case with the reported PDE coefficients,
        integrated over a longer time horizon than in training (i.e, $T=50$),
        superimposed on the test-set initial codes (gray).}
    \label{fig:cavia-latent-space-adr}
\end{figure}

\begin{figure}[ht!]
    \centering
    \includegraphics[scale=1.]{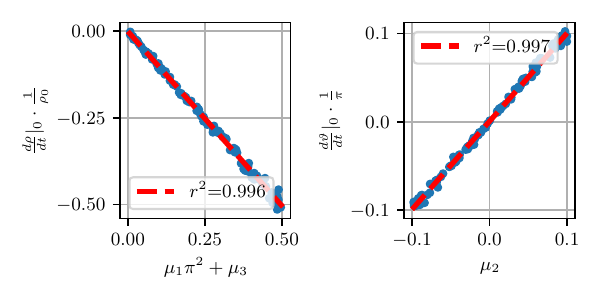}
    \caption{ADR test case: initial latent velocity against PDE parameters (Meta-IC-LDNet).
    For each test-set sample, the initial latent velocity is expressed in polar coordinates $(\rho, \vartheta)$
        with respect to the center of the initial codes.
        The radial velocity, normalized by the initial radius $\rho_0$, is plotted against $\mu_1\pi^2+\mu_3$ (left),
        and the angular velocity, normalized by $\pi$, against $\mu_2$ (right).
        The red dashed lines are linear fits, with the corresponding coefficient of determination $r^2$.}
    \label{fig:cavia-latent-adr-params}
\end{figure}

As before, the 2D latent space can be conveniently visualized. Specifically,
Figure~\ref{subfig:cavia-initial-latent-adr}
shows the initial test-set latent states, colored according to the amplitude
and phase of the corresponding initial conditions. Interestingly, compared to
Figure~\ref{fig:autodec-initial-latent-adr}, the latent space shows the desired topology, with the phase periodically distributed. This suggests that meta-learning encourages a more structured organization of the
latent space, avoiding performance degradation during long roll-outs. In fact, when considering advection-dominated evolutions in the physical space, the closed latent space topology corresponds to trajectories that remain within the training region, where the decoder performs the best, as visualized in Figure~\ref{subfig:cavia-advection-adr}.
Finally, in Figure~\ref{fig:cavia-latent-adr-params}, we depict a nearly perfect correlation between the PDE coefficients and the initial velocity in
the latent space, using polar coordinates $\rho$ and $\vartheta$.
As observed in Figure~\ref{fig:autodec-latent-traj-adr}, the diffusion and reaction coefficients influence the movement in the amplitude
direction, while the advection coefficient affects the motion in the phase direction, consistently with the physical
interpretation of ADR.

\begin{figure}[p]
    \centering
    \begin{subfigure}{0.9\textwidth}
        \centering
        \includegraphics[scale=1.]{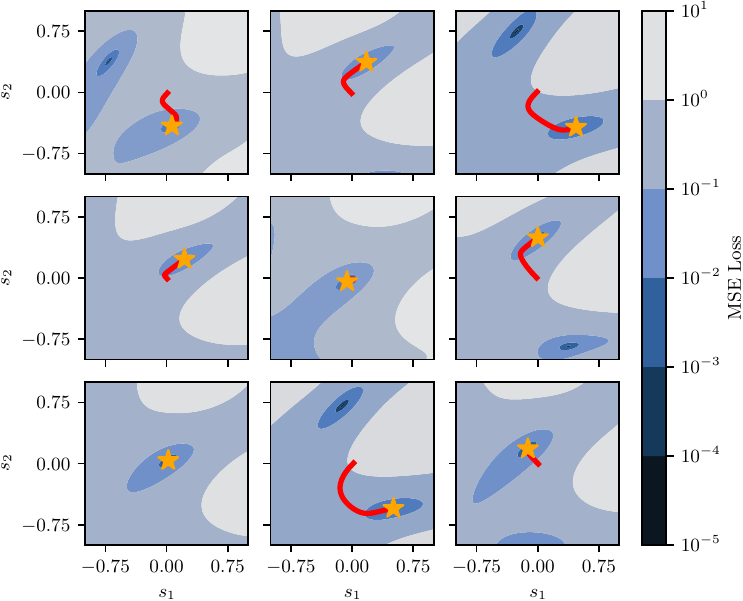}
        \caption{AD-IC-LDNet}
        \label{subfig:autodec-latent-fitting-adr}
    \end{subfigure}
    \begin{subfigure}{0.9\textwidth}
        \centering
        \includegraphics[scale=1.]{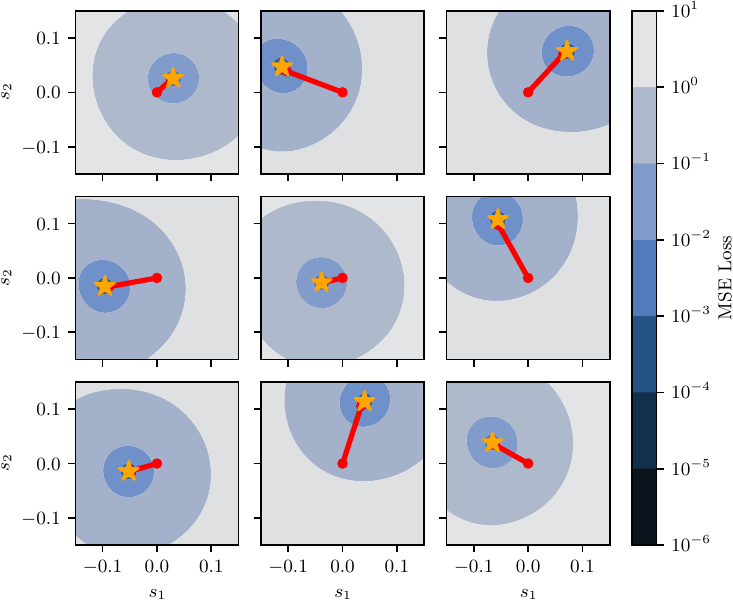}
        \caption{Meta-IC-LDNet}
        \label{subfig:cavia-latent-fitting-adr}
    \end{subfigure}
    \caption{ADR test case: initial-code fitting on the test set.
    Each panel refers to one of the first nine test-set samples
        (same in (a) and (b))
        and shows the contours of the MSE loss on the initial condition
        as a function of the initial latent code $(s_1, s_2)$.
        The red line is the optimization path and the orange star is the final code
        (i.e., $10^4$ Adam iterations from a random initialization close to the origin for the AD-IC-LDNet,
        and three gradient-descent steps from $\mathbf{s}_0 = \mathbf{0}$ for the Meta-IC-LDNet).}
    \label{fig:latent-fitting-adr}
\end{figure}

\subsubsection{AD-IC-LDNet versus Meta-IC-LDNet: initial-code fitting on the ADR test case}
\label{subsubsec:adr-latent-fitting-comparison}
Both the AD-IC-LDNet and the Meta-IC-LDNet share the same approach of
treating the initial latent codes as learnable vectors (i.e., auto-decoding). However, at training time, the former learns them together with the model parameters,
while the latter adopts a two-loop approach. At inference time, both strategies require an optimization procedure to find the
optimal initial codes for the unseen samples. Since $d_s=2$, the loss function and the optimization
trajectory can be conveniently visualized. In particular, Figure~\ref{fig:latent-fitting-adr} shows the loss-function contours for both the AD-IC-LDNet and the Meta-IC-LDNet, along with the optimization paths to fit the initial codes for
the first nine test-set examples.
Comparing Figures~\ref{subfig:autodec-latent-fitting-adr} and~\ref{subfig:cavia-latent-fitting-adr},
we observe that meta-learning favors a better conditioned and more regular loss. Indeed, the circular level curves correspond to a well-conditioned Hessian (i.e., similar maximum and minimum eigenvalues) near the minimum, guaranteeing a rapid convergence of gradient-based optimization~\cite{bottou_optimization_2018}. In contrast, the
AD-IC-LDNet exhibits an irregular landscape, with multiple local minima.
These differences are the result of forcing the Meta-IC-LDNet to learn new tasks in only three optimization steps,
which requires a well-conditioned landscape.
Thus, the Meta-IC-LDNet is significantly more efficient in adapting to new tasks, allowing for rapid online inference.

\begin{figure}[hb!]
    \centering
    \begin{subfigure}{0.49\textwidth}
        \centering
        \includegraphics[scale=1.]{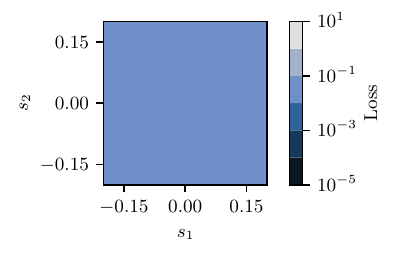}
        \caption{10\,000 epochs}
        \label{subfig:10k-latent-space-adr}
    \end{subfigure}
    \begin{subfigure}{0.49\textwidth}
        \centering
        \includegraphics[scale=1.]{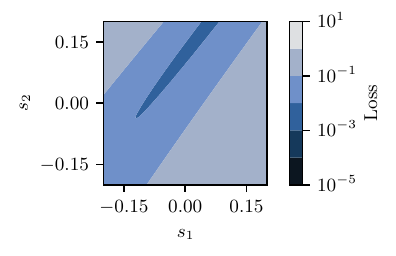}
        \caption{15\,000 epochs}
        \label{subfig:15k-latent-space-adr}
    \end{subfigure}
    \begin{subfigure}{0.49\textwidth}
        \centering
        \includegraphics[scale=1.]{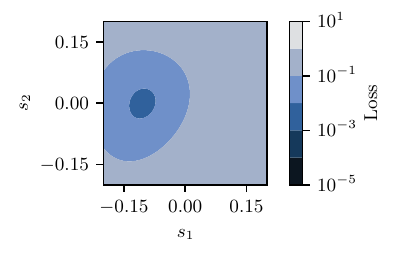}
        \caption{19\,000 epochs}
        \label{subfig:19k-latent-space-adr}
    \end{subfigure}
    \begin{subfigure}{0.49\textwidth}
        \centering
        \includegraphics[scale=1.]{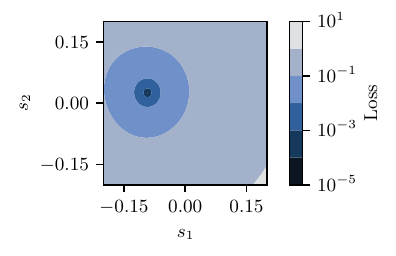}
        \caption{45\,000 epochs}
        \label{subfig:45k-latent-space-adr}
    \end{subfigure}
    \caption{ADR test case: evolution of the inner-loop loss function (Meta-IC-LDNet).
    Contours of the inner-loop loss as a function of the initial latent code $(s_1, s_2)$,
        for the same validation-set sample at selected training epochs.
        The corresponding training and validation losses are shown in Figure~\ref{fig:train_val_adr_cavia}.}
    \label{fig:cavia-latent-space-evolution-adr}
\end{figure}

It is also interesting to analyze how the inner-loop loss function of the Meta-IC-LDNet evolves during training
to observe how it achieves the final well-conditioned and convex landscape.
In particular, Figure~\ref{fig:cavia-latent-space-evolution-adr} plots the inner-loop loss function at selected
epochs, for the same validation-set sample,
and should be read together with Figure~\ref{fig:train_val_adr_cavia} (see Appendix~\ref{sec:appendix-train-val}).
After an initial phase where the loss is essentially flat, it develops into a narrow valley, reflecting an ill-conditioned Hessian. As training progresses, between 17\,000 and 18\,000 epochs, the
condition number of the Hessian suddenly improves, leading to the final regular and smooth loss landscape.
At the same time, the minimum value of the loss function continues to decrease. From preliminary experiments, we observed that this sudden improvement of the loss landscape is a crucial step to achieve good training
performance and generalization, and the Meta-IC-LDNet appears to be very sensitive to the choice of hyperparameters. Hence, fast adaptation and correct latent-space topology come at the expense of robustness and ease of training.

%% file: Sections/result_subsec/cylinder.tex
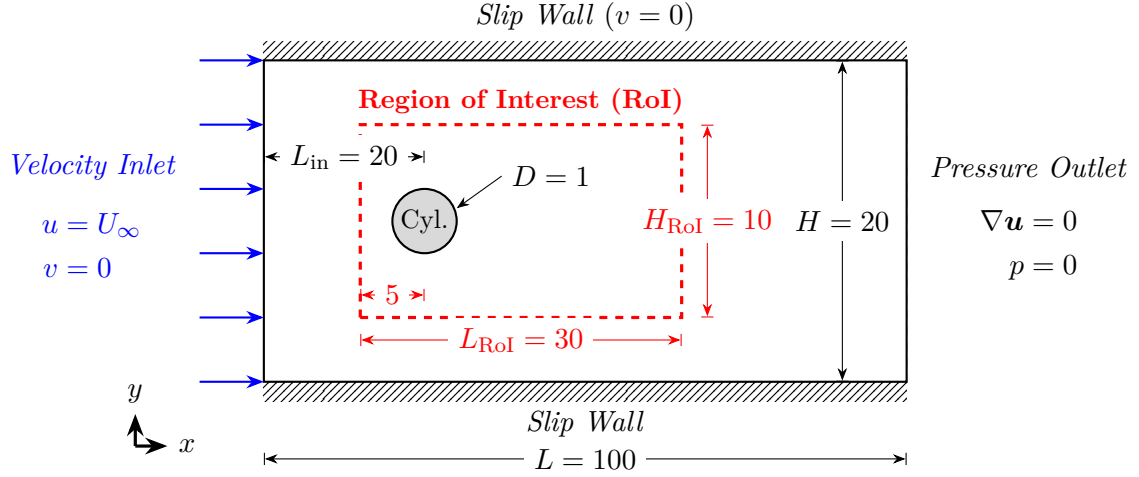
\begin{figure}[t!]
    \centering
    \begin{tikzpicture}[>=Stealth, scale=0.85]

        % --- PARAMETERS ---
        \def\L{10}       
        \def\H{5}        
        \def\cx{2.5}       
        \def\cy{2.5}       
        \def\D{1}        
        \def\R{\D/2}     

        % --- RoI PARAMETERS ---
        \def\boxOffsetXp{4.}
        \def\boxOffsetXn{1}
        \def\boxOffsetY{1.5}

        % --- DOMAIN ---
        \draw[thick] (0,0) rectangle (\L,\H);

        \draw[thick, fill=gray!30] (\cx,\cy) circle (\R);
        \node at (\cx,\cy) {\small Cyl.};

        % --- RoI ---
        \draw[red, very thick, dashed]
        (\cx - \boxOffsetXn, \cy - \boxOffsetY) rectangle (\cx + \boxOffsetXp, \cy + \boxOffsetY);

        \node[red, above, font=\bfseries\small] at (\cx-\boxOffsetXn+\boxOffsetXp*0.5+\boxOffsetXn*0.5, \cy +
        \boxOffsetY) {Region of Interest (RoI)};

        % --- BC ---

        % 1. INLET
        \foreach \y in {0, 1., ..., \H} {
            \draw[->, thick, blue] (-1, \y) -- (0, \y);
        }
        \node[anchor=east, blue, align=center] at (-1.2, \H/2) {
            \emph{Velocity Inlet} \\[2 ex]
            $
            \begin{aligned}
                u &= U_\infty \\
                v &= 0
            \end{aligned}
            $
        };

        % 2. OUTLET 
        \draw[dashed] (\L,0) -- (\L,\H);
        \node[anchor=west, align=center] at (\L+0.2, \H/2) {
            \emph{Pressure Outlet} \\[2 ex]
            $
            \begin{aligned}
                \nabla \boldsymbol{u} &= 0 \\
                p &= 0
            \end{aligned}
            $
        };

        % 3. WALLS
        \fill[pattern=north east lines] (0, \H) rectangle (\L, \H+0.3);
        \fill[pattern=north east lines] (0, 0) rectangle (\L, -0.3);
        \node[above] at (\L/2, \H+0.3) {\emph{Slip Wall} $( v = 0 )$};
        \node[below] at (\L/2, -0.3) {\emph{Slip Wall} };

        % --- DIMENSIONS ---

        % L
        \draw[|<->|] (0, -1.2) -- (\L, -1.2) node[midway, fill=white] {$L=100$};

        % H
        \draw[|<->|] (\L-1., 0) -- (\L-1., \H) node[midway, fill=white] {$H=20$};

        % L RoI
        \draw[|<->|, red] (\cx - \boxOffsetXn, \cy - \boxOffsetY - 0.35) -- (\cx + \boxOffsetXp, \cy - \boxOffsetY
        - 0.35) node[red, midway, fill=white] {$L_{\mathrm{RoI}}=30$};

        \draw[|<->|, red] (\cx - \boxOffsetXn, \cy - \boxOffsetY + 0.35) -- (\cx, \cy - \boxOffsetY +
        0.35) node[red, midway, fill=white] {$5$};

        % H RoI
        \draw[|<->|, red] (\cx + \boxOffsetXp + 0.4, \cy - \boxOffsetY) -- (\cx + \boxOffsetXp + 0.4, \cy + \boxOffsetY
        ) node[red, midway, fill=white] {$H_{\mathrm{RoI}}=10$};

        % inlet distance
        \draw[|<->|] (0, \cy+\R+0.5) -- (\cx, \cy+\R+0.5) node[midway, fill=white] {$L_{\mathrm{in}}=20$};

        % D
        \draw[<-] (\cx+2*\R, \cy) ++(-0.5,0.2) -- ++(0.7, 0.5) node[right] {$D=1$};

        % --- AXES ---
        \draw[->, very thick] (-2.,-1.) -- (-1.5, -1.) node[right] {$x$};
        \draw[->, very thick] (-2.,-1.) -- (-2., -0.5) node[above] {$y$};

    \end{tikzpicture}
    \caption{2D flow around a circular cylinder: domain and boundary conditions (not to scale). The dataset fields are sampled in the region of interest (RoI, red dashed box).}
    \label{fig:cyl-setup}

\end{figure}
The second test case covers the 2D flow around a static circular cylinder for a Reynolds number $\mathrm{Re}_D=100$,
corresponding to the regime of laminar vortex shedding~\cite{lienhard_synopsis_1966}.
The computational domain and boundary conditions are
schematically represented in Figure~\ref{fig:cyl-setup}.
Specifically, the domain is a rectangle of length $L=100$ and height $H=20$,
with a circular-cylinder cutout of diameter $D=1$
placed at a distance $L_{\mathrm{in}}=20$ from the inlet and centered vertically. The top and bottom boundaries are
treated as slip walls, while a uniform velocity profile $U_\infty$ is imposed at the
inlet, ramping up from zero over one second of simulation time. At the outlet, the pressure is set to zero, together
with a zero-gradient condition on the velocity field. Finally, the
cylinder
surface is treated as a no-slip wall.
The fluid properties are set to $\nu=1$, resulting in $Re_D=U_{\infty}$.
The case is simulated with OpenFOAM \texttt{v2506}, using a structured mesh with approximately one million cells,
with progressive refinement boxes to ensure adequate resolution around the cylinder and in its proximal wake region.
The simulation is run using the \texttt{pimpleFoam} laminar-flow solver, for a
total of four seconds of simulation time,
with an adaptive time step to maintain a Courant number $\mathrm{Co} \leq 1.0$.
\begin{figure}[b!]
    \centering
    \begin{subfigure}{0.49\textwidth}
        \centering
        \includegraphics[scale=1.]{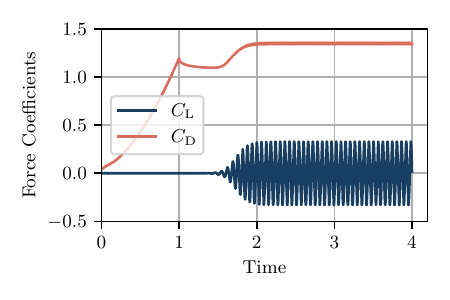}
        \caption{Force coefficients in time.}
        \label{subfig:cyl-force-coeffs-time}
    \end{subfigure}
    \begin{subfigure}{0.49\textwidth}
        \centering
        \includegraphics[scale=1.]{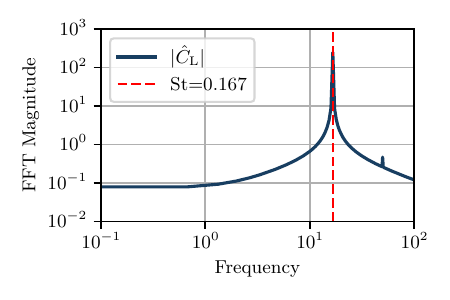}
        \caption{Lift-coefficient FFT.}
        \label{subfig:cl-fft}
    \end{subfigure}
    \caption{2D flow around a circular cylinder: force coefficients from the OpenFOAM simulation
    at $\mathrm{Re}_D=100$. (a) Time history of the lift and drag coefficients. (b)
        FFT magnitude of the lift coefficient for the developed flow. The dashed line identifies the shedding frequency, corresponding to $\mathrm{St}=0.167$.}
    \label{fig:cyl-force-coeffs}
\end{figure}
Figure~\ref{subfig:cyl-force-coeffs-time} shows the time history of the drag and lift coefficients,
highlighting the periodic vortex shedding behavior, while Figure~\ref{subfig:cl-fft} shows the
FFT of the lift coefficient for the fully developed flow.
The numerical Strouhal number is approximately $\mathrm{St}=0.167$, in good
agreement
with experimental values~\cite{lienhard_synopsis_1966}.
Finally, in order to build the deep-learning dataset, the pressure and velocity fields are sampled every $10^{-3}$ seconds of simulation time, on a uniform $600 \times 200$ structured grid of points,
covering the red-dashed box in Figure~\ref{fig:cyl-setup}.
In all cases, the flow variables are extracted from the value at the nearest cell center.
Since the flow has a period of approximately 60 time steps, the
dataset consists of 60 overlapping sequences, starting from consecutive moments within the shedding period, each of length
61 time steps. The dataset is randomly split into training, validation, and test sets, with an 80\%-10\%-10\% ratio.
Since each training sequence spans a full shedding period, every test initial condition also appears as an intermediate state of some training trajectory. This test case therefore primarily assesses the identification of the correct latent state from a single snapshot, whereas generalization to unseen states is assessed in the other test cases. Additionally, in this test case, $\mathbf{u}(t)$ is absent, because the dynamical system is autonomous.

\begin{figure}[p]
    \centering
    \begin{subfigure}{\textwidth}
        \centering
        \makebox[\linewidth][c]{\includegraphics[scale=1.]{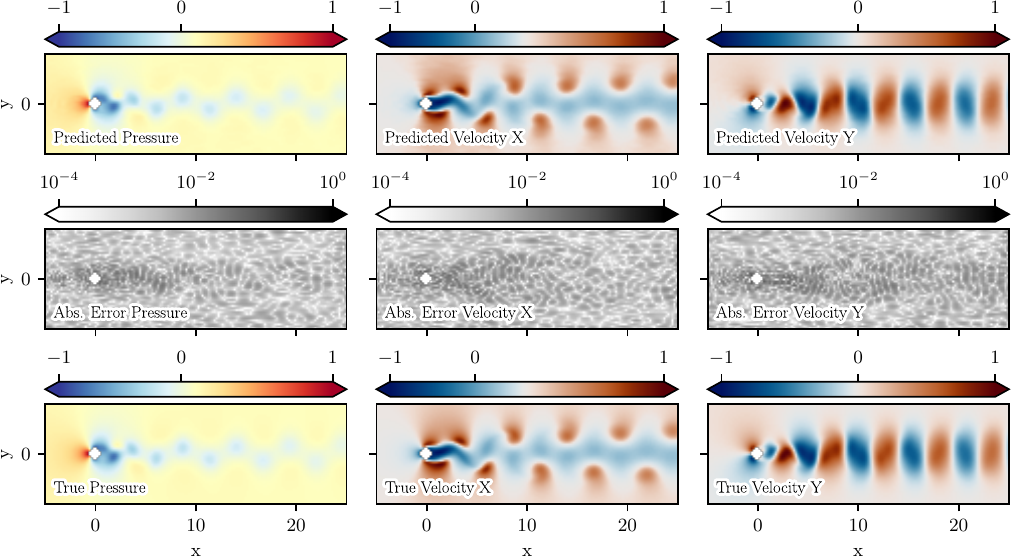}}
        \caption{Test-set sample 1}
        \label{subfig:cyl100-autodec-fields-sample-1}
    \end{subfigure}
    \begin{subfigure}{\textwidth}
        \centering
        \makebox[\linewidth][c]{\includegraphics[scale=1.]{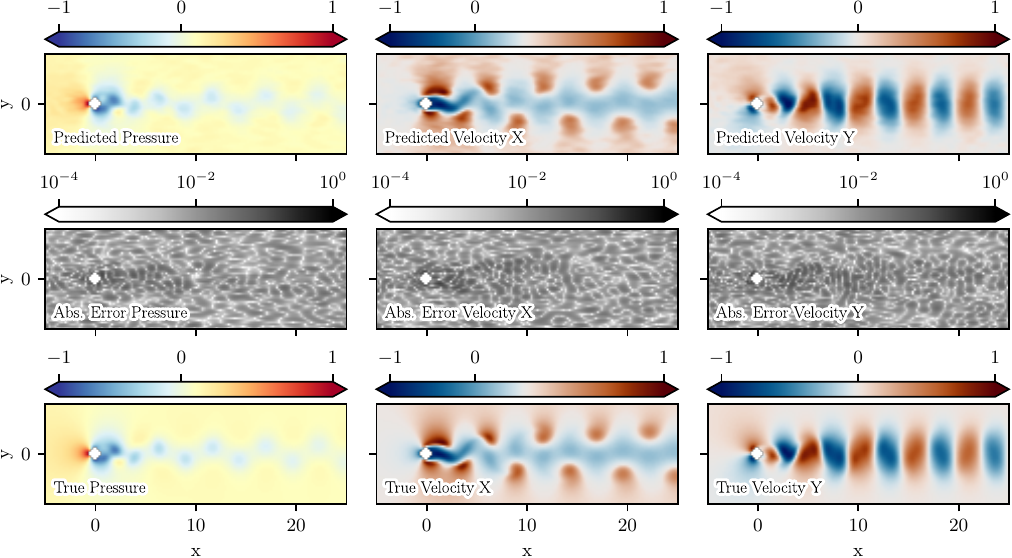}}
        \caption{Test-set sample 3}
        \label{subfig:cyl100-autodec-fields-sample-3}
    \end{subfigure}
    \caption{Static-cylinder test case: final-time normalized test-set fields (AD-IC-LDNet).
    Predicted (top), absolute error (middle), and true (bottom) pressure,
        $x$-velocity, and $y$-velocity at the last time step of test-set samples 1 (a) and 3 (b),
        numbered as in Figure~\ref{fig:autodec-initial-latent-cylinder}.
        The fields are normalized to $[-1, 1]$ using the training-set minimum and maximum values.}
    \label{fig:cyl100-autodec-test-fields-final}
\end{figure}
\begin{figure}[b!]
    \centering
    \begin{subfigure}{0.49\textwidth}
        \centering
        \includegraphics[scale=1.]{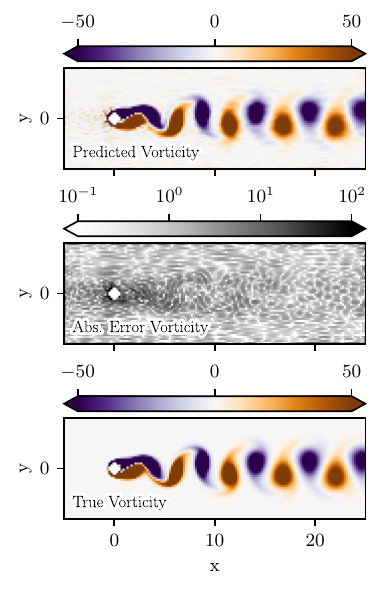}
        \caption{Test-set sample 1}
        \label{subfig:cyl100-autodec-omega-sample-1}
    \end{subfigure}
    \begin{subfigure}{0.49\textwidth}
        \centering
        \includegraphics[scale=1.]{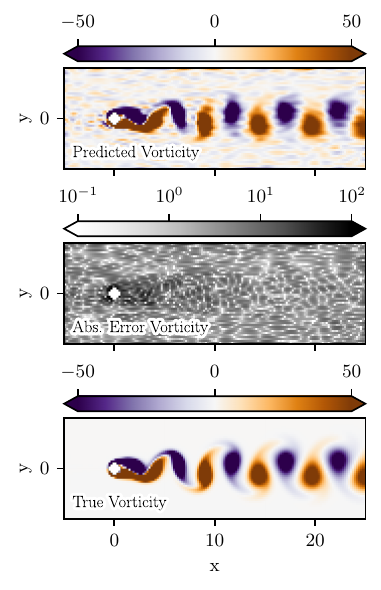}
        \caption{Test-set sample 3}
        \label{subfig:cyl100-autodec-omega-sample-3}
    \end{subfigure}
    \caption{Static-cylinder test case: final-time test-set vorticity (AD-IC-LDNet).
    Predicted (top), absolute error (middle), and true (bottom) vorticity at the last time step of test-set samples 1 (a) and 3 (b).
    The predicted vorticity is obtained by automatic differentiation of the predicted velocity field, while the true one is computed by OpenFOAM.}
    \label{fig:cyl100-autodec-test-omega-final}
\end{figure}

\subsubsection{AD-IC-LDNet on the static-cylinder test case}
\label{subsubsec:autodec-cyl}
On the cylinder-flow dataset, the AD-IC-LDNet is trained to predict the
evolution of pressure and
velocity fields from an initial snapshot. To reduce the computational burden,
the original $200 \times 600$ grid of the region of interest is downscaled to a
coarser $50 \times 150$ resolution. As summarized in Table~\ref{tab:hyperparameters-comparison-cyl} (see Appendix~\ref{sec:appendix-hyper}) the architecture consists of a fully connected $\mathcal{NN}_{\mathrm{dyn}}$
and a FiLM-modulated~\cite{perez_film_2018} SIREN-based~\cite{sitzmann_implicit_2020} $\mathcal{NN}_{\mathrm{rec}}$. Specifically, the modulation approach consists of a phase shift of the
sinusoidal activations, similar to that proposed in~\cite{du_conditional_2024}, and is summarized as follows:
\begin{equation}
    \left\{
        \begin{aligned}
            \boldsymbol{h}_{0} &= \boldsymbol{x}(t) \\
            \boldsymbol{h}_{i} &= \sin\left( \omega_{i} \cdot \big( \boldsymbol{W}_{i} \boldsymbol{h}_{i-1} + \boldsymbol{b}_{i} +
            \boldsymbol{W}^{\mathrm{cond}}_{i} \boldsymbol{s}(t) + \boldsymbol{b}^{\mathrm{cond}}_i \big) \right)
            \quad \forall i = 1, \dots, N_{\mathrm{hidden}}^{\mathrm{dec}} \\
            \tilde{\boldsymbol{y}} &= \boldsymbol{W}_{\mathrm{out}} \boldsymbol{h}_{N_{\mathrm{hidden}}^{\mathrm{dec}}} + \boldsymbol{b}_{
                \mathrm{out}}
        \end{aligned}
    \right.
\end{equation}
where $\boldsymbol{h}_i$ is the output of the $i$-th hidden layer and $\omega_i$ is a frequency hyperparameter (set to
30 for every layer). $\boldsymbol{W}_i$ and $\boldsymbol{b}_i$ are the weights and biases of the $i$-th hidden layer, while $\boldsymbol{W}^{\mathrm{cond}}_i$
and $\boldsymbol{b}^{\mathrm{cond}}_i$ are the weights and biases of the conditioning branch, which takes as input the latent state
$\boldsymbol{s}(t)$ and produces a modulation signal for the activations.
Note that, although the original SIREN features $\omega_i$ only in the first layer, our implementation is equivalent in
terms of initialization.
Due to the essentially periodic nature of the flow, similar to a simple harmonic oscillator, we expect
two latent variables to be necessary and sufficient to capture the main dynamics. The two-dimensional state should provide a representation equivalent to the one that considers the phase and the
angular velocity of the vortex shedding. Hence, the latent space dimension is set to $d_s=2$.
\begin{figure}[ht!]
    \centering
    \includegraphics[scale=1.]{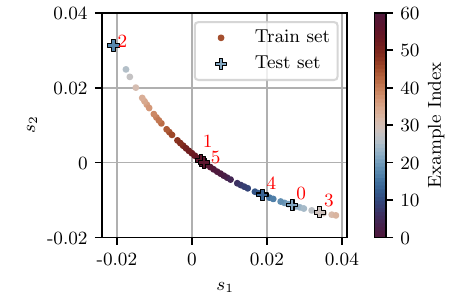}
    \caption{Static-cylinder test case: initial latent states (AD-IC-LDNet).
    Training-set initial codes, colored by the index of the initial time step within the shedding period,
        and test-set initial codes (crosses),
        numbered for reference in Figures~\ref{fig:cyl100-autodec-test-fields-final} and~\ref{fig:cyl100-autodec-test-omega-final}.}
    \label{fig:autodec-initial-latent-cylinder}
\end{figure}

The trained model is evaluated on the test set and Figure~\ref{fig:cyl100-autodec-test-fields-final}
shows
the
predicted fields for two selected test-set samples at the final time step.
Note that the training-set
minimum and
maximum
values are used to linearly normalize the data.
The first
example (i.e., Figure~\ref{subfig:cyl100-autodec-fields-sample-1}) exhibits a smaller final error, while the second (i.e., Figure~\ref{subfig:cyl100-autodec-fields-sample-3}) shows a more noticeable degradation of the surrogate solution.
In particular, the error is dominated by components characterized by a high spatial frequency.
Overall, the average NRMSE across the test dataset is $4.1 \times 10^{-3}$ for the pressure channel and $4.7 \times 10^{-3}$ and $7.1 \times 10^{-3}$ for the two velocity components, respectively. 
Furthermore, since the IC-LDNet is a neural field, we can compute derived quantities, such as the vorticity field,
by means of automatic differentiation. In particular, Figure~\ref{fig:cyl100-autodec-test-omega-final}
shows the predicted vorticity field for the same two test-set samples at the final time step,
where the high-frequency error is even more evident, as the differentiation amplifies it. 

Interestingly, the unevenly distributed error can be explained by the suboptimal latent-space structure produced by AD-IC-LDNet.
In particular,
Figure~\ref{fig:autodec-initial-latent-cylinder} shows the initial codes for the training and test sets,
colored according to the time index of the initial condition within the shedding period.
For clarity, the test codes are also uniquely identified by a number.
We observe that the initial codes are arranged
in a one-dimensional manifold, where the position along the manifold is related to the time within the shedding period.
Hence, although the flow is periodic, the neural surrogate produces open trajectories in the latent space,
leading to degraded predictive performance outside the training region. That explains why, in Figures~\ref{fig:cyl100-autodec-test-fields-final} and~\ref{fig:cyl100-autodec-test-omega-final}, the predictions for sample 3 were significantly less accurate than those for sample 1.

\subsubsection{Meta-IC-LDNet on the static-cylinder test case}
\label{subsubsec:cavia-cyl}
The Meta-IC-LDNet is also trained on the cylinder-flow dataset
with the hyperparameters summarized in Table~\ref{tab:hyperparameters-comparison-cyl} (see Appendix~\ref{sec:appendix-hyper}). In this case, the inner loss function only covers the
initial condition (i.e., $T_{\mathrm{inner}}^{\mathrm{train}}=0$), while the outer loss is computed on the entire time sequence (i.e., $\mathcal{T}_{\mathrm{outer}}^i = [0, T]$).

\begin{figure}[p]
    \centering
    \captionsetup{skip=6pt}
    \includegraphics[scale=1.]{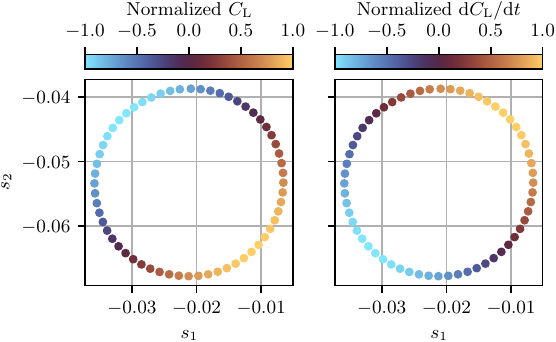}
    \caption{Static-cylinder test case: physical interpretation of initial latent states (Meta-IC-LDNet).
    Initial codes of the training, validation, and test sets combined,
        colored by the lift coefficient $C_{\mathrm{L}}$ (left) and its time derivative $\mathrm{d}C_{\mathrm{L}}/\mathrm{d}t$ (right) at the initial time, both min-max normalized to $[-1, 1]$.}
    \label{fig:cavia-initial-latent-cylinder-physics}
    \vspace{\floatsep}
    \makebox[\linewidth][c]{\includegraphics[scale=1.]{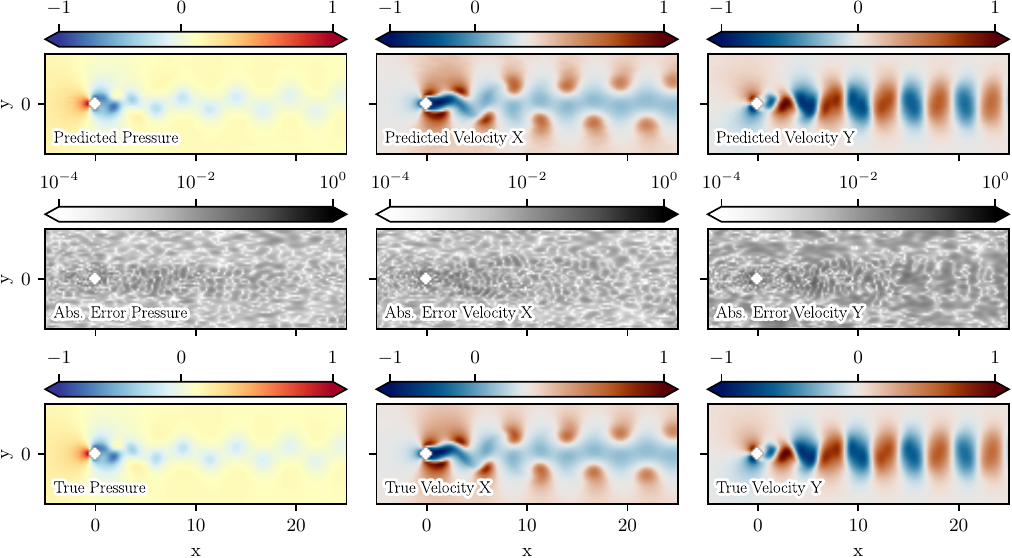}}
    \caption{Static-cylinder test case: final-time normalized test-set fields (Meta-IC-LDNet).
        Predicted (top), absolute error (middle), and true (bottom)
        normalized fields at the last time step of test-set sample 1,
        arranged as in Figure~\ref{fig:cyl100-autodec-test-fields-final}.}
    \label{fig:cavia-fields-cylinder}
    \vspace{\floatsep}
    \makebox[\linewidth][c]{\includegraphics[scale=1.]{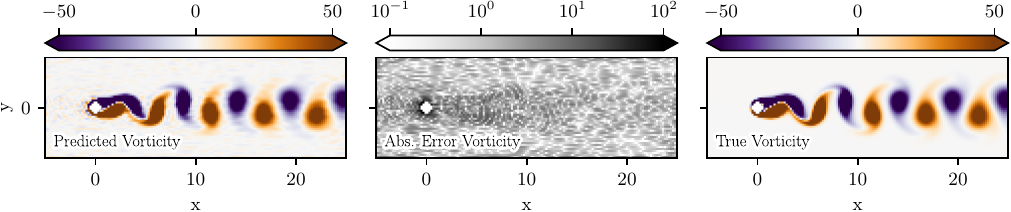}}
    \caption{Static-cylinder test case: final-time test-set vorticity (Meta-IC-LDNet).
    Predicted (left), absolute error (center), and true (right) vorticity at the last time step of test-set sample 1,
        computed as in Figure~\ref{fig:cyl100-autodec-test-omega-final}.}
    \label{fig:cavia-vorticity-cylinder}
\end{figure}
Tuning experiments showed sudden instabilities during the training process, which culminated in the
model failing to train properly. Hence, we employed the curriculum-learning strategy described
in Section~\ref{subsubsec:regularization-stability}.
In particular, the outer loss is computed on the first two time steps for 500 epochs, then a new time step is
added every 100 epochs, eventually covering the entire sequence of 61 time steps. Hence, the model
first learns the initial latent codes and velocities, which, for this test case, are crucial to capture the periodic behavior.
Indeed, as shown in Figure~\ref{fig:cavia-initial-latent-cylinder-physics}, this approach leads to a latent space
with the expected topology, where the codes are arranged in a closed manifold,
revealing a correspondence with $C_{\mathrm{L}}$ and $\mathrm{d}C_{\mathrm{L}} / \mathrm{d}t$, and therefore with the angular position and velocity of the shedding, respectively.

The correct topology and the physical adherence of the latent space allow for long roll-outs
without loss of accuracy, as the latent trajectory is closed and remains within the training region.
In this regard, Figure~\ref{fig:cavia-fields-cylinder} shows the predicted fields for test-set sample 1 at the last time step, demonstrating the stability over time.
The topologically consistent latent-space also leads to an even distribution of the prediction error across the test set, with all the examples characterized by a similar NRMSE of around $2.6 \times 10^{-3}$, $2.7 \times 10^{-3}$, and $4.1 \times 10^{-3}$, for pressure and velocity components, respectively.
Furthermore, Figure~\ref{fig:cavia-vorticity-cylinder} shows the predicted vorticity for the same sample and time step, highlighting the ability of the model to accurately capture the whirling structures.
Hence, these results demonstrate that the Meta-IC-LDNet, with the help of curriculum learning,
trains a surrogate model that captures the periodicity and the physics of the flow,
leading to accurate and stable predictions.

%% file: Sections/result_subsec/randomrot.tex
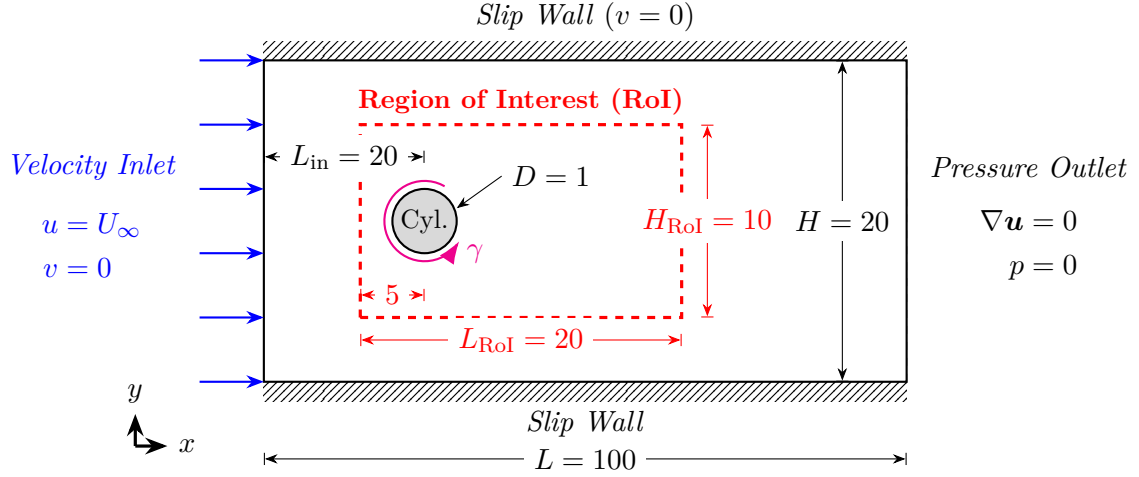
\begin{figure}[t!]
    \centering
    \begin{tikzpicture}[>=Stealth, scale=0.85]

        % --- PARAMETERS ---
        \def\L{10}       
        \def\H{5}        
        \def\cx{2.5}       
        \def\cy{2.5}       
        \def\D{1}        
        \def\R{\D/2}     

        % --- RoI PARAMS ---
        \def\boxOffsetXp{4.}
        \def\boxOffsetXn{1}
        \def\boxOffsetY{1.5}

        % --- DOMAIN ---
        \draw[thick] (0,0) rectangle (\L,\H);

        \draw[thick, fill=gray!30] (\cx,\cy) circle (\R);
        \node at (\cx,\cy) {\small Cyl.};

        \draw[magenta, thick, -{Latex[length=3mm]}] plot[domain=60:330, samples=40] ({1.25*\R*cos(\x) + \cx}, {1.25*\R*sin(\x) + \cy});
        \node[magenta, anchor=west] at (\cx+0.5, \cy - 0.5) {$\gamma$};

        % --- RoI ---
        \draw[red, very thick, dashed]
        (\cx - \boxOffsetXn, \cy - \boxOffsetY) rectangle (\cx + \boxOffsetXp, \cy + \boxOffsetY);

        \node[red, above, font=\bfseries\small] at (\cx-\boxOffsetXn+\boxOffsetXp*0.5+\boxOffsetXn*0.5, \cy +
        \boxOffsetY) {Region of Interest (RoI)};

        % --- BCs ---

        % 1. INLET 
        \foreach \y in {0, 1., ..., \H} {
            \draw[->, thick, blue] (-1, \y) -- (0, \y);
        }
        \node[anchor=east, blue, align=center] at (-1.2, \H/2) {
            \emph{Velocity Inlet} \\[2 ex]
            $
            \begin{aligned}
                u &= U_\infty \\
                v &= 0
            \end{aligned}
            $
        };

        % 2. OUTLET
        \draw[dashed] (\L,0) -- (\L,\H);
        \node[anchor=west, align=center] at (\L+0.2, \H/2) {
            \emph{Pressure Outlet} \\[2 ex]
            $
            \begin{aligned}
                \nabla \boldsymbol{u} &= 0 \\
                p &= 0
            \end{aligned}
            $
        };

        % 3. WALLS 
        \fill[pattern=north east lines] (0, \H) rectangle (\L, \H+0.3);
        \fill[pattern=north east lines] (0, 0) rectangle (\L, -0.3);
        \node[above] at (\L/2, \H+0.3) {\emph{Slip Wall} $( v = 0 )$};
        \node[below] at (\L/2, -0.3) {\emph{Slip Wall} };

        % --- DIMENSIONS ---

        % L
        \draw[|<->|] (0, -1.2) -- (\L, -1.2) node[midway, fill=white] {$L=100$};

        % H
        \draw[|<->|] (\L-1., 0) -- (\L-1., \H) node[midway, fill=white] {$H=20$};

        % L roi
        \draw[|<->|, red] (\cx - \boxOffsetXn, \cy - \boxOffsetY - 0.35) -- (\cx + \boxOffsetXp, \cy - \boxOffsetY
        - 0.35) node[red, midway, fill=white] {$L_{\mathrm{RoI}}=20$};

        \draw[|<->|, red] (\cx - \boxOffsetXn, \cy - \boxOffsetY + 0.35) -- (\cx, \cy - \boxOffsetY +
        0.35) node[red, midway, fill=white] {$5$};

        % H roi
        \draw[|<->|, red] (\cx + \boxOffsetXp + 0.4, \cy - \boxOffsetY) -- (\cx + \boxOffsetXp + 0.4, \cy + \boxOffsetY
        ) node[red, midway, fill=white] {$H_{\mathrm{RoI}}=10$};

        % Inlet
        \draw[|<->|] (0, \cy+\R+0.5) -- (\cx, \cy+\R+0.5) node[midway, fill=white] {$L_{\mathrm{in}}=20$};

        % D
        \draw[<-] (\cx+2*\R, \cy) ++(-0.5,0.2) -- ++(0.7, 0.5) node[right] {$D=1$};

        % --- AXES ---
        \draw[->, very thick] (-2.,-1.) -- (-1.5, -1.) node[right] {$x$};
        \draw[->, very thick] (-2.,-1.) -- (-2., -0.5) node[above] {$y$};

    \end{tikzpicture}
    \caption{2D flow around a rotating circular cylinder: domain and boundary conditions (not to scale).
    Same configuration as in Figure~\ref{fig:cyl-setup}, with the cylinder rotating at a time-varying angular velocity $\gamma$ and a smaller region of interest.}
    \label{fig:randomrot-setup}

\end{figure}
\begin{figure}[t!]
    \centering
    \includegraphics[scale=1.]{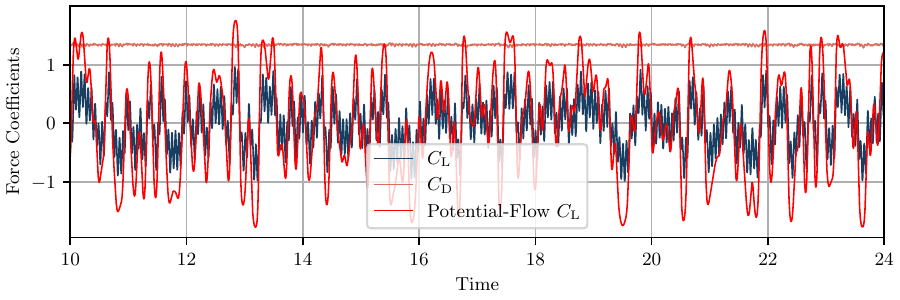}
    \caption{2D flow around a rotating circular cylinder: force coefficients.
    Lift and drag coefficients from the OpenFOAM simulation over the time window used to build the dataset,
        together with the angular velocity of the cylinder.
        To match the scale of the lift and drag coefficients, the angular velocity is reported as the equivalent potential-flow $C_{\mathrm{L}}$.}
    \label{fig:force-coefficients-randomrot}
\end{figure}
Following the test on the surrogate modeling of the flow around a
static circular cylinder, the IC-LDNet is further assessed on a more challenging
extension of the same problem, where the cylinder is rotating with an angular velocity $\gamma$ that varies randomly in time. To build the dataset,
a similar OpenFOAM setup to the
one described in Section~\ref{subsec:test-case-cyl} is considered,
with the only difference being that a moving-wall boundary condition is applied to the cylinder
wall, as summarized in Figure~\ref{fig:randomrot-setup}. Here, the region of interest only covers the
$[-5, 15] \times [-5, 5]$ domain (with a resolution of
$400 \times 200$) to reduce the memory footprint of the stored snapshots. The
simulation is run for
24 seconds of physical time, with an adaptive time step to maintain a Courant number below 1.0, and the snapshots
are stored every $2\times10^{-3}$ seconds. To develop the flow, the cylinder is initially stationary for five seconds, and then starts rotating at a random angular velocity. This kinematic profile is generated by sampling a zero-mean Gaussian process (squared-exponential kernel, correlation length $0.05$ s), subsequently bounded within $\pm 60$ rad/s using a tanh filter. 

To assemble the final dataset, only the snapshots corresponding to the last 14 seconds of simulation time are considered,
resulting in a total of 7\,000 snapshots. Specifically, Figure~\ref{fig:force-coefficients-randomrot} summarizes the
evolution of the force coefficients over time, alongside the corresponding angular velocity of the cylinder, representing the input $\mathbf{u}(t)$.
The simulation is partitioned into 70 non-overlapping sequences of 100 snapshots each, and the resulting dataset is
randomly split into
training,
validation, and test sets,
in a
ratio of 80\%, 10\%, and 10\%, respectively.

\begin{figure}[t!]
    \centering
    \begin{subfigure}{0.99\textwidth}
        \centering
        \includegraphics[scale=1.]{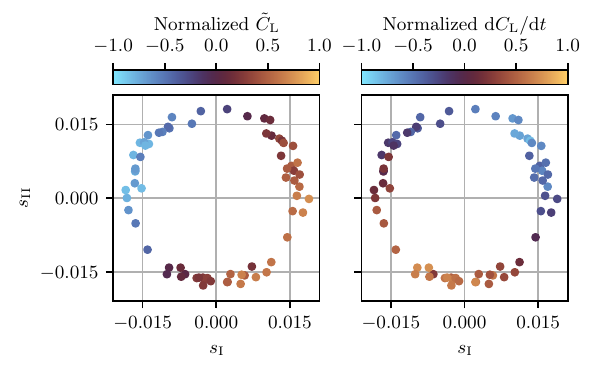}
        \caption{$s_{\mathrm{I}}$ and $s_{\mathrm{II}}$}
        \label{subfig:randomrot-cavia-codes-principal-12}
    \end{subfigure}

    \begin{subfigure}{0.99\textwidth}
        \centering
        \includegraphics[scale=1.]{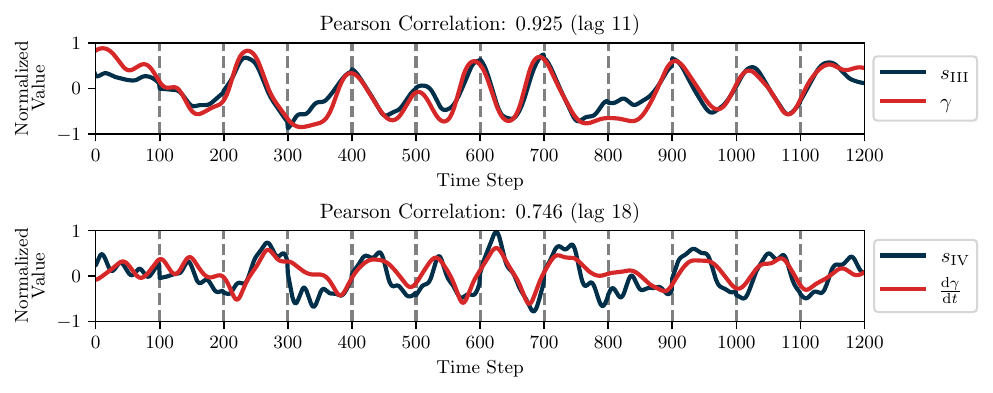}
        \caption{$s_{\mathrm{III}}$ and $s_{\mathrm{IV}}$}
        \label{subfig:randomrot-cavia-codes-principal-34}
    \end{subfigure}
    \caption{Rotating-cylinder test case: latent-space PCA (Meta-IC-LDNet).
    The principal components $s_{\mathrm{I}}, \dots, s_{\mathrm{IV}}$
        are computed on the latent trajectories of the training, validation, and test sets combined.
        (a) Initial codes, colored by $\tilde{C}_{\mathrm{L}}$ (i.e.,
        the lift coefficient minus its sliding average over 30 time steps) and by $\mathrm{d}C_{\mathrm{L}}/\mathrm{d}t$.
        (b) Third and fourth principal components along twelve consecutive sequences (separated by dashed lines),
        compared with the angular velocity $\gamma$ and its time derivative,
        delayed by the lag (in time steps) that maximizes the Pearson correlation.
        The colors in (a) and the curves in (b) are min-max normalized to $[-1, 1]$.}
    \label{fig:cavia-latent-randomrot-pca}
\end{figure}
An instance of the Meta-IC-LDNet, whose hyperparameters are summarized in Table~\ref{tab:hyperparameters-rotating-cylinder} (see Appendix~\ref{sec:appendix-hyper}), is trained on the rotating-cylinder dataset.
Unlike Test Case 2A, the dataset is not downsized to a fixed spatial resolution, but the loss function is computed on 1\,250
randomly sampled points from the region of interest, instead of the full-resolution grid (i.e., 80\,000 points).
Every snapshot within each time series is spatially subsampled independently. At test time, the approximate solution is instead computed at 40\,000 randomly sampled points, shared across samples and time steps.
Moreover, as before, the curriculum-learning strategy
described in Section~\ref{subsubsec:regularization-stability} is applied, gradually exposing the model to the entire simulations.

\begin{figure}[tb!]
    \centering
    \includegraphics[scale=1.]{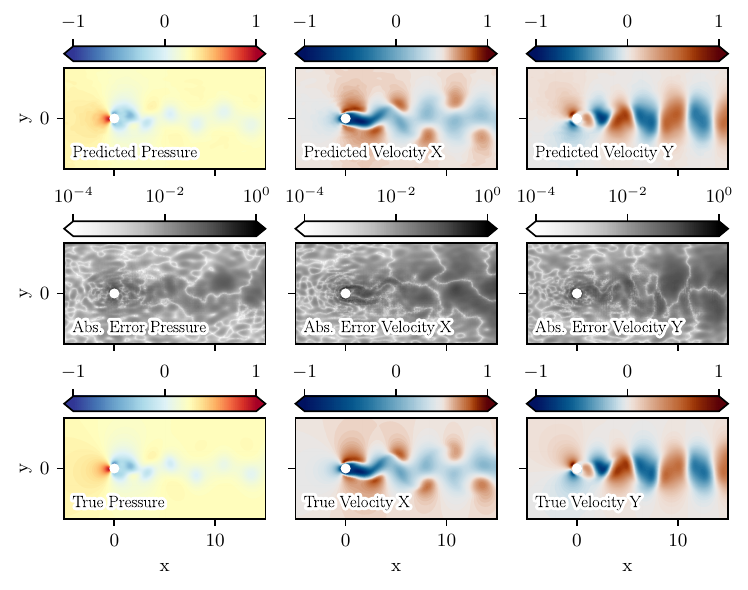}
    \caption{Rotating-cylinder test case: final-time normalized test-set fields (Meta-IC-LDNet).
    Predicted (top), absolute error (middle), and true (bottom) normalized fields at the last time step of the first test-set sample,
        arranged as in Figure~\ref{fig:cyl100-autodec-test-fields-final}.}
    \label{fig:cavia-fields-randomrot}
    \vspace{\floatsep}
    \includegraphics[scale=1.]{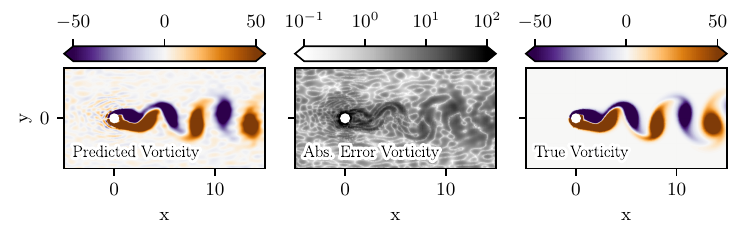}
    \caption{Rotating-cylinder test case: final-time test-set vorticity (Meta-IC-LDNet).
    Predicted (left), absolute error (center), and true (right) vorticity at the last time step of the first test-set sample,
        computed as in Figure~\ref{fig:cyl100-autodec-test-omega-final}.}
    \label{fig:cavia-vorticity-randomrot}
\end{figure}
In line with the static-cylinder findings, the latent space learned by the Meta-IC-LDNet exhibits a topologically consistent arrangement of the initial codes, and a physically meaningful
relationship is observed between the latent codes and the flow features. 
To better analyze the structure of the latent space, whose dimension is here $d_s=4$, Principal Component Analysis (PCA) is performed on the latent trajectories, considering
the combined training,
validation, and test sets. The first two principal components of the initial codes are shown in
Figure~\ref{subfig:randomrot-cavia-codes-principal-12},
while Figure~\ref{subfig:randomrot-cavia-codes-principal-34} depicts a subset of latent trajectories
projected onto the third and fourth principal components. The former plot suggests that the first two principal
components are related to the high-frequency periodic shedding, similar to the observations in
Figure~\ref{fig:cavia-initial-latent-cylinder-physics}. In particular, $\tilde{C}_{\mathrm{L}}$ represents the
difference between the instantaneous lift coefficient and a sliding average of $C_{\mathrm{L}}$ over a time window
of 30 time steps (i.e., approximately one shedding period), highlighting high-frequency variations. Additionally, Figure~\ref{subfig:randomrot-cavia-codes-principal-34}
reveals a strong correlation of the third and fourth principal components with the cylinder angular velocity and its
time derivative, respectively, provided that a certain time shift is considered (i.e., the one that maximizes the correlation).
This suggests that the latent space contains past information on the cylinder rotation.
Hence, the four-dimensional latent
space encodes both the high-frequency von Kármán shedding and the
low-frequency Magnus effect~\cite{anderson_fundamentals_2024}.

As observed in previous tests, this well-organized latent space allows
for good generalization and an accurate prediction of the flow evolution. For example,
Figures~\ref{fig:cavia-fields-randomrot} and~\ref{fig:cavia-vorticity-randomrot} show the true and predicted flow fields and vorticity
contours at the last time step of the first test-set example, along with the corresponding absolute error. As before, the
vorticity is computed from the predicted velocity field through automatic differentiation.
Although the
error is noticeably larger than the one observed in the static-cylinder case, the model is still able to reconstruct the
flow field with a good degree of accuracy, despite the more challenging nature of the problem.
Overall, on the test set, the pressure is predicted with an average NRMSE of $7.5 \times 10^{-3}$, while the x and y velocity predictions are characterized by an average NRMSE of $1.2 \times 10^{-2}$ and $1.3 \times 10^{-2}$, respectively. 

\begin{figure}[ht!]
    \centering
    \begin{subfigure}{0.5\textwidth}
        \centering
        \includegraphics[scale=1.]{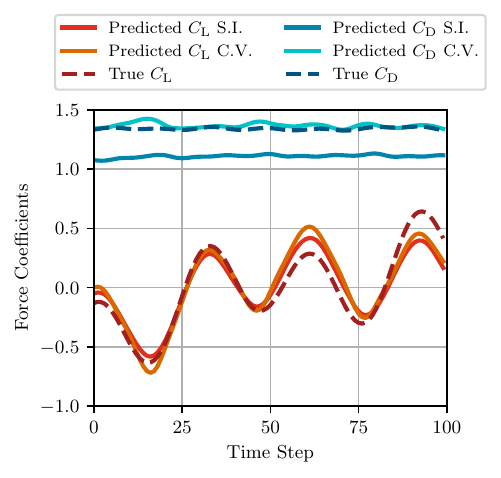}
        \caption{Original}
        \label{subfig:randomrot-cavia-force-coeffs-orig}
    \end{subfigure}%
    \begin{subfigure}{0.5\textwidth}
        \centering
        \includegraphics[scale=1.]{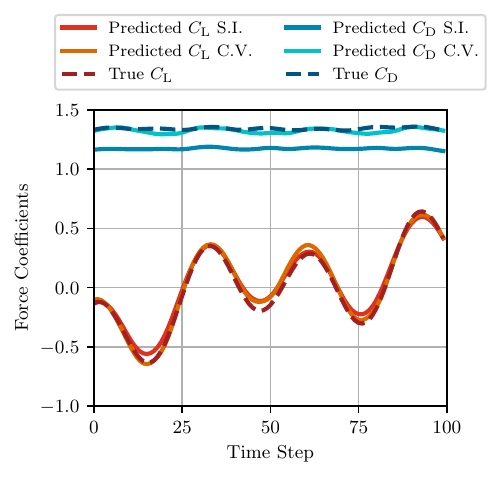}
        \caption{Fine-tuned}
        \label{subfig:randomrot-cavia-force-coeffs-finetune}
    \end{subfigure}
    \caption{Rotating-cylinder test case: force coefficients (Meta-IC-LDNet).
    Lift and drag coefficients of the first test-set sample, computed from the predicted fields via surface integration
        (S.I.) on the cylinder wall and via momentum balance on a control volume (C.V.) around the cylinder,
        and compared with the OpenFOAM reference (dashed lines).
        (a) Original model. (b) Model fine-tuned on a denser grid of points,
        including an O-grid around the cylinder.}
    \label{fig:randomrot-cavia-force-coeffs}
\end{figure}

Finally, the approximate velocity and pressure fields are used to compute the force coefficients,
which are compared to the ones computed in OpenFOAM.
Figure~\ref{fig:randomrot-cavia-force-coeffs} shows the time evolution of the lift and drag coefficients for the
first test-set example. Specifically, two methods for computing the force coefficients are considered: surface integration (i.e., S.I.)
of the pressure and viscous forces on the cylinder wall, and momentum balance on a fluid control volume (i.e., C.V.) surrounding the cylinder.
Interestingly, while both methods provide similarly satisfactory results for the lift coefficient,
the control-volume approach is highly accurate for the drag coefficient, which is underestimated by the
surface-integration method, possibly due to the training-set resolution near the cylinder wall.
Indeed, in Figure~\ref{subfig:randomrot-cavia-force-coeffs-finetune}, when considering an improved model fine-tuned for an additional 24 hours on a denser grid, the accuracy of both lift and drag coefficients considerably increases. Specifically, the fine-tuning grid consists of 40\,000 randomly sampled points from the RoI (see Figure~\ref{fig:randomrot-setup}), as well as 14\,500 additional randomly sampled points from a $58 \times 500$ (radial $\times$ angular) O-grid covering the annular region $0.48\leq r \leq 1$, where $r$ denotes the distance from the cylinder center. Note that the O-grid slightly extends inside the cylinder to better capture the boundary.

\begin{figure}[t!]
    \centering
    \begin{subfigure}{0.49\textwidth}
        \centering
        \includegraphics[scale=1.]{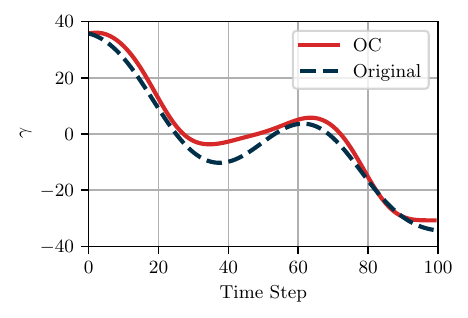}
        \caption{Optimal-control rotation.}
        \label{subfig:randomrot-optimal-control-signal}
    \end{subfigure}
    \begin{subfigure}{0.49\textwidth}
        \centering
        \includegraphics[scale=1.]{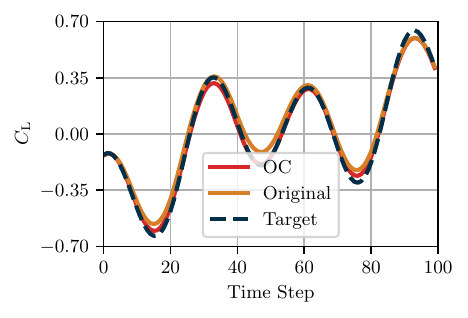}
        \caption{Optimal-control lift coefficient.}
        \label{subfig:randomrot-optimal-control-cl}
    \end{subfigure}
    \caption{Rotating-cylinder test case: optimal control (Meta-IC-LDNet).
    The fine-tuned model is used to recover the angular velocity that reproduces the
    lift coefficient of the first test-set sample.
        (a) Angular velocity obtained by solving the optimal-control (OC) problem, compared with the original one.
        (b) Lift coefficient predicted by the surrogate model with the OC and with the original angular velocity,
        compared with the OpenFOAM target.}
    \label{fig:randomrot-optimal-control}
\end{figure}

Accurate surrogate predictions of the force coefficients pave the way for the application of
the Meta-IC-LDNet to optimal control (OC) problems, which require many evaluations of the flow field. As a proof
of concept, the fine-tuned model is tasked with recovering the angular velocity that produces the lift-coefficient evolution of the
first test-set example (i.e., the one in Figure~\ref{fig:randomrot-cavia-force-coeffs}) by minimizing
the MSE loss between the predicted (via the S.I. method) and target $C_{\mathrm{L}}$ over the entire time series.
The optimization is performed in $2\times 10^3$ Adam iterations, using a learning rate of $10^{-3}$ and $L_2$ regularization.
In Figure~\ref{subfig:randomrot-optimal-control-signal}, the original and recovered angular velocities are compared. Although
the error between the two signals is not negligible, the optimal-control angular velocity follows the same trend as the original. Furthermore, in Figure~\ref{subfig:randomrot-optimal-control-cl} we compare the target lift coefficient (i.e., from OpenFOAM) to the one predicted by the surrogate model using the original rotation and the one using the just computed optimal-control rotation. Specifically, Figure~\ref{subfig:randomrot-optimal-control-cl} shows that the discrepancy between the original and reconstructed signal in Figure~\ref{subfig:randomrot-optimal-control-signal} does not originate from the solution of the optimal-control problem, since the OC $C_{\mathrm{L}}$ accurately follows the target.
Crucially, the fast inference of the neural surrogate allows for solving the optimal-control problem in approximately 16 seconds, highlighting
the potential of the proposed framework for real-world many-query applications.

%% file: Sections/result_subsec/beam.tex
\begin{figure}[t!]
    \centering
    \tdplotsetmaincoords{65}{115}
    \begin{tikzpicture}[tdplot_main_coords, line join=round, line cap=round, scale=1.2]

    % --- PARAMS ---
        \def\LInc{1.25}      
        \def\LFace{0.75}     
        \def\LBeam{6.}     
        \def\HQx{1.}
        \def\HQy{1.}
        \def\HQz{1.}
        \def\RQ{0.5}

    % --- 1. REFERENCE AXES ---
    \begin{scope}[shift={(1.,1.,-3.)}]
        \draw[-{Latex[length=3mm]}, thick] (0,0,0) -- (1.,0,0) node[anchor=north east] {$x$};
        \draw[-{Latex[length=3mm]}, thick] (0,0,0) -- (0,0.75,0) node[anchor=north west] {$y$};
        \draw[-{Latex[length=3mm]}, thick] (0,0,0) -- (0,0,0.75) node[anchor=south] {$z$};
    \end{scope}

    % --- 2. CLAMP ---
    \fill[gray!15] (0,-\LInc,-\LInc) -- (0,\LInc,-\LInc) -- (0,\LInc,\LInc) -- (0,-\LInc,\LInc) -- cycle;
    \fill[pattern=north east lines, pattern color=gray!50] (0,-\LInc,-\LInc) -- (0,\LInc,-\LInc) -- (0,\LInc,\LInc) -- (0,-\LInc,\LInc) -- cycle;
    \draw[thick, black!60] (0,-\LInc,-\LInc) -- (0,\LInc,-\LInc) -- (0,\LInc,\LInc) -- (0,-\LInc,\LInc) -- cycle;

    % --- 3. FACES ---
    % RIGHT
    \fill[black!10] (0,\LFace,-\LFace) -- (\LBeam,\LFace,-\LFace) -- (\LBeam,\LFace,\LFace) -- (0,\LFace,\LFace) -- cycle;
    % TOP
    \fill[black!5] (0,-\LFace,\LFace) -- (\LBeam,-\LFace,\LFace) -- (\LBeam,\LFace,\LFace) -- (0,\LFace,\LFace) -- cycle;
    % FREE
    \fill[black!20] (\LBeam,-\LFace,-\LFace) -- (\LBeam,\LFace,-\LFace) -- (\LBeam,\LFace,\LFace) -- (\LBeam,-\LFace,\LFace) -- cycle;

    % --- 4. EDGES ---
    \draw[thick] (\LBeam,-\LFace,-\LFace) -- (\LBeam,\LFace,-\LFace) -- (\LBeam,\LFace,\LFace) -- (\LBeam,-\LFace,\LFace) -- cycle; 
    \draw[thick] (0,\LFace,-\LFace) -- (\LBeam,\LFace,-\LFace);
    \draw[thick] (0,\LFace,\LFace) -- (\LBeam,\LFace,\LFace); 
    \draw[thick] (0,-\LFace,\LFace) -- (\LBeam,-\LFace,\LFace); 
    \draw[thick] (0,\LFace,-\LFace) -- (0,\LFace,\LFace) -- (0,-\LFace,\LFace);

    % --- 5. LOADS ---

    % a) AXIAL
    \foreach \y in {-\LFace, \LFace} {
        \foreach \z in {-\LFace, \LFace} {
            \draw[-{Latex[length=2.5mm]}, red, thick] (\LBeam, \y, \z) -- (\LBeam + \HQx, \y, \z);
        }
    }
    \draw[red, thick, dashed] (\LBeam + \HQx, -\LFace, -\LFace) -- (\LBeam + \HQx, \LFace, -\LFace) -- (\LBeam + \HQx, \LFace, \LFace) -- (\LBeam + \HQx, -\LFace, \LFace) -- cycle;
    \fill[red, opacity=0.1] (\LBeam + \HQx, -\LFace, -\LFace) -- (\LBeam + \HQx, \LFace, -\LFace) -- (\LBeam + \HQx, \LFace, \LFace) -- (\LBeam + \HQx, -\LFace, \LFace) -- cycle;
    \node[red, anchor=south west] at (\LBeam + 2*\HQx, -1.3*\LFace, 0.3*\LFace) {$\mathbf{f}_x$};

    % b) Y SHEAR
    \foreach \y in {\LFace} {
        \foreach \z in {-\LFace, \LFace} {
            \draw[-{Latex[length=2.5mm]}, green!60!black, thick] (\LBeam, \y, \z) -- (\LBeam, \y+\HQy, \z);
        }
    }
    \draw[green!60!black, thick, dashed] (\LBeam, \LFace, -\LFace) -- (\LBeam, \LFace + \HQy, -\LFace) -- (\LBeam, \LFace + \HQy, \LFace) -- (\LBeam, \LFace, \LFace) -- cycle;
    \fill[green!60!black, opacity=0.1] (\LBeam, \LFace, -\LFace) -- (\LBeam, \LFace + \HQy, -\LFace) -- (\LBeam, \LFace + \HQy, \LFace) -- (\LBeam, \LFace, \LFace) -- cycle;
    \node[green!60!black, anchor=south] at (\LBeam, \LFace + 1.3*\HQy, -0.5*\LFace) {$\mathbf{f}_y$};

    % c) Z SHEAR
    \foreach \y in {-\LFace, \LFace} {
        \foreach \z in {\LFace} {
            \draw[-{Latex[length=2.5mm]}, blue, thick] (\LBeam, \y, \z) -- (\LBeam, \y, \z+\HQz);
        }
    }
    \draw[blue, thick, dashed] (\LBeam, -\LFace, \LFace + \HQz) -- (\LBeam, \LFace, \LFace + \HQz) -- (\LBeam, \LFace, \LFace) -- (\LBeam, -\LFace, \LFace) -- cycle;
    \fill[blue, opacity=0.1] (\LBeam, -\LFace, \LFace + \HQz) -- (\LBeam, \LFace, \LFace + \HQz) -- (\LBeam, \LFace, \LFace) -- (\LBeam, -\LFace, \LFace) -- cycle;
    \node[blue, anchor=north] at (\LBeam, -1.5*\LFace, \LFace + 0.9*\HQz) {$\mathbf{f}_z$};

    % d) TWISTING COUPLE
    \draw[magenta, thick, -{Latex[length=3mm]}] plot[domain=60:270, samples=40] ({\LBeam}, {\RQ*cos(\x)}, {\RQ*sin(\x)});
    \node[magenta, anchor=west] at (\LBeam, 0.7*\RQ, 0) {$\mathbf{q}$};

    \end{tikzpicture}
    \caption{Nonlinear solid dynamics of a 3D cantilever beam: domain and boundary conditions (not to scale).
    The beam is clamped at $x=0$, while axial ($\mathbf{f}_x$) and shear ($\mathbf{f}_y$, $\mathbf{f}_z$) tractions and a twisting moment ($\mathbf{q}$) are applied on the free end.}
    \label{fig:beam-setup}
\end{figure}
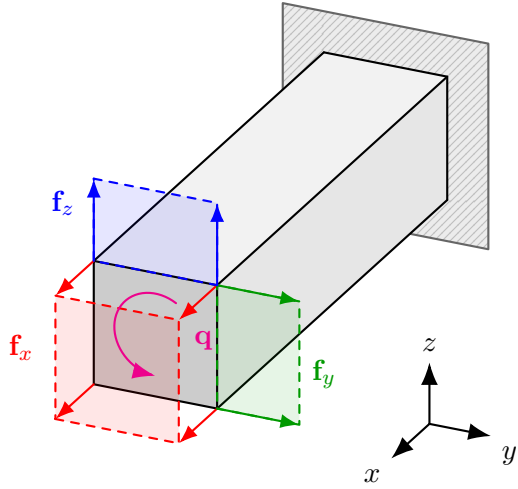

The final test case considered in this work deals with the nonlinear solid dynamics of a 3D cantilever beam.
As summarized in Figure~\ref{fig:beam-setup}, the beam is clamped at one end and loads are applied on the free extremity. Specifically,
the instantaneous intensities of these loads are randomly sampled
using independent zero-mean Gaussian processes. All processes share a temporal correlation length of $l = 1.0$, 
with a standard deviation set to $\sigma_f = 0.05$ for the axial and shear tractions, and $\sigma_q = 500$ for the twisting moment intensity.

The beam has a length of $L = 0.1$ and a square cross-section with side length $a = 0.01$,
and is made of a hyperelastic material modeled by a
compressible Neo-Hookean constitutive law~\cite{holzapfel_nonlinear_2000}. Specifically, the strain energy density $\varPsi$ is described as:
\begin{equation}
    \label{eq:neo-hook-psi}
    \varPsi = \frac{\mu}{2} \left( J^{-2/3} I_1 - 3 \right) + \frac{\lambda}{4} \left( (J - 1)^2 + (\ln J)^2 \right),
\end{equation}
where $\lambda$ is the first Lamé parameter, $\mu$ is the shear modulus (i.e., the second Lamé parameter), $J=\det(\mathbf{F})$ with $\mathbf{F}$ being the deformation gradient, and $I_1=\mathrm{tr}(\mathbf{C})$ with $\mathbf{C} = \mathbf{F}^T \mathbf{F}$ being the right Cauchy-Green tensor.
The density of the material is $\rho = 10^3$ and the damping coefficient
is $\eta = 10^2$. To introduce variability in the material properties, the shear modulus is sampled from a uniform distribution in the range $[25, 250]$, while the first Lamé parameter is proportionally updated as $\lambda = 2.5\: \mu$. Note that we are considering non-dimensional physical quantities.

The dynamic evolution of the system is governed by the Principle of Virtual Work. Specifically, expressed in a total Lagrangian framework over the reference configuration $\Omega$, the problem consists in finding a kinematically admissible displacement field $\mathbf{d}$ such that, for any valid test function $\mathbf{v}$, the following equation holds:
\begin{equation}
    \label{eq:weak-form}
    \int_{\Omega} \rho \, \ddot{\mathbf{d}} \cdot \mathbf{v} \,\mathrm{d}\Omega + 
    \int_{\Omega} \left(
    \mathbf{P}  + \frac{\eta}{2} (\mathbf{F} \dot{\mathbf{C}}) \right): \nabla_0 \mathbf{v} \, \mathrm{d}\Omega = \int_{\Gamma_N} \left( \mathbf{f}_t + q_t \boldsymbol{\tau} \right) \cdot \mathbf{v} \, \mathrm{d}S,
\end{equation}
where $\mathbf{P} = \partial \varPsi / \partial \mathbf{F}$ is the first Piola-Kirchhoff stress tensor, $\nabla_0$ denotes the gradient operator with respect to the material coordinates, $\ddot{\mathbf{d}}$ is the acceleration field, and $\dot{\mathbf{C}}$ is the rate of the right Cauchy-Green tensor accounting for viscous dissipation. The external work on the Neumann boundary $\Gamma_N$ (i.e., the free extremity) depends on the applied surface tractions $\mathbf{f}_t$ and the twisting moment controlled by $q_t$ through the lever-arm vector $\boldsymbol{\tau} = [0, Z, -Y]^T$.

\begin{figure}[b!]
    \centering
    \begin{subfigure}{0.99\textwidth}
        \centering
        \includegraphics[scale=1.]{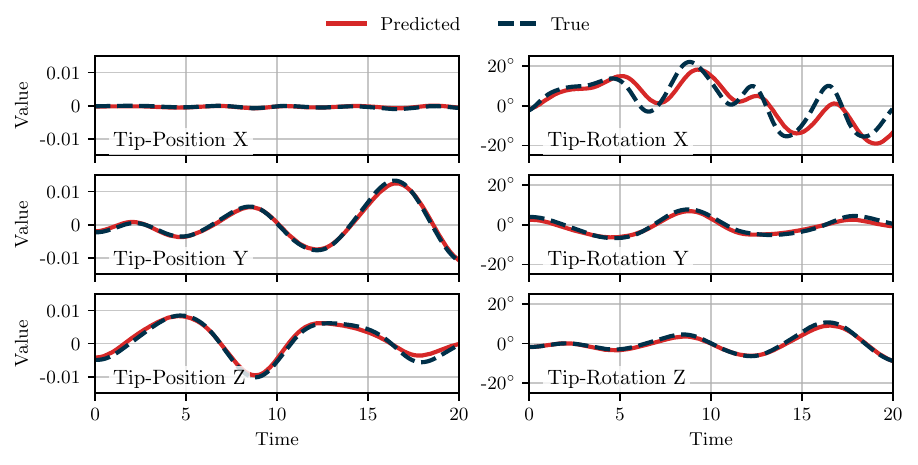}
        \caption{Best test-set prediction: tip dynamics.}
        \label{subfig:cavia_best_pred_beam}
    \end{subfigure}
    \begin{subfigure}{0.99\textwidth}
        \centering
        \includegraphics[scale=1.]{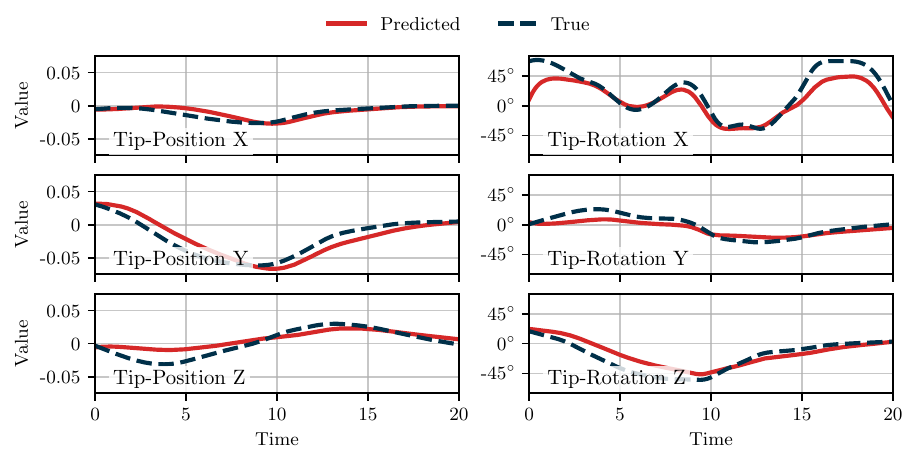}
        \caption{Worst test-set prediction: tip dynamics.}
        \label{subfig:cavia_worst_pred_beam}
    \end{subfigure}
    \caption{Nonlinear-beam test case: inference on the test set (Meta-IC-LDNet).
    Predicted and true displacement at the center of the free-end cross-section (left)
        and rotation angles of the cross-section about the $x$, $y$, and $z$ axes, computed with the Kabsch algorithm (right),
        for the test-set samples with the lowest (a) and highest (b) NRMSE.}
    \label{fig:cavia-inference-beam}
\end{figure}

The spatial domain is discretized using a structured mesh of first-order tetrahedral elements with a characteristic side length of $2.5 \times 10^{-3}$. Time integration is performed using a fully implicit finite difference scheme, approximating the acceleration field and the kinematic rates with backward differences. The resulting fully discrete nonlinear system is solved at each time step $t_n$ by means of the Newton-Raphson method, implemented within the FEniCS finite element framework~\cite{logg_automated_2012,alnaes_fenics_2015}. The simulation spans a total of 2\,500 time steps with a constant time increment of $\Delta t = 2 \times 10^{-2}$. Initially, at $t=0$, the system starts from rest, assuming zero displacement and velocity fields.

To compose the dataset, 500 unique configurations of load histories and material properties are simulated, and the nodal
displacements are recorded at each time step. Then, for each simulation, the first 500 time steps are discarded,
and the remaining 2\,000 are split into two non-overlapping sequences of 1\,000 time steps each, resulting in a total of 1\,000 sequences. The input $\mathbf{u}(t)$ contains both the constant shear modulus and the variable intensities of the external loads.
The final dataset is randomly partitioned into training, validation, and test sets with a split of
80\%, 10\%, and 10\%, respectively.

The Meta-IC-LDNet is trained as a surrogate for the FEniCS solver on the nonlinear beam dynamics.
Unlike the previous test cases, the model is trained on a workstation equipped with an NVIDIA GeForce RTX 5090 GPU and an Intel Core Ultra 9 285K CPU with 64 GB of RAM.
In this case, $\mathcal{NN}_{\mathrm{dyn}}$ and $\mathcal{NN}_{\mathrm{rec}}$ are both FCNNs with SiLU activation function, and the latent space has a dimension of $d_s=12$ (refer to Table~\ref{tab:hyperparameters-nonlinear-beam} for the full set of hyperparameters). To alleviate the memory burden, the original 1\,025 points of the mesh are randomly subsampled to
500 points during training, while the full set of points is used during inference.
Finally, the second regularization strategy described in
Section~\ref{subsubsec:regularization-stability} is adopted.
Specifically, three additional inner-loop optimizations are performed at equidistant times within the sequence,
and the MSE between
the resulting codes and the corresponding state in the latent trajectory is added to the loss function.

\begin{figure}[t!]
    \centering
    \includegraphics[scale=1.]{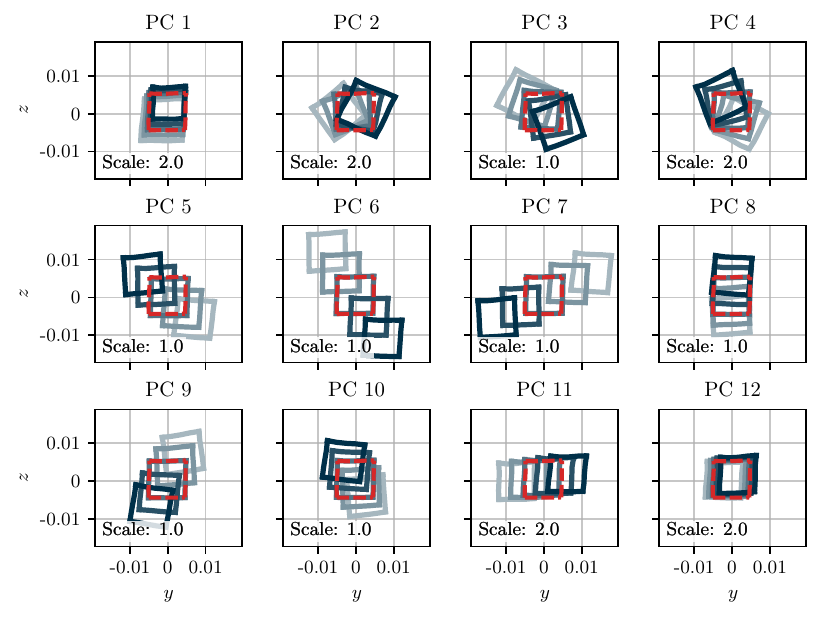}
    \caption{Nonlinear-beam test case: effect of latent-space principal components (Meta-IC-LDNet).
    Each panel shows the free-end cross-section in the $yz$ plane,
        decoded from latent states where a single principal component (PC)
        varies between its 1st and 99th percentiles over all latent states (from light gray to dark blue),
        while the remaining components are fixed at their mean values.
        The red dashed square corresponds to all components at their mean values.
        Displacements are amplified by the indicated scale factor for visualization.}
    \label{fig:effect-pc-beam-cavia-section}
\end{figure}

The performance and generalization of the trained model are assessed on the test set. Specifically,
Figure~\ref{fig:cavia-inference-beam} presents the tip dynamics of the best and worst predictions within the test set.
The position of the tip is extracted directly from the model outputs (i.e., the displacement components),
while the beam-section rotation at the free extremity is computed during post-processing using the Kabsch algorithm~\cite{lawrence_purely_2019}.
In particular, Figure~\ref{subfig:cavia_best_pred_beam} shows that the model can accurately predict the beam dynamics
with only a slight discrepancy in torsion.
By contrast, in the worst-case scenario (i.e., Figure~\ref{subfig:cavia_worst_pred_beam}),
although the surrogate predictions exhibit a noticeable error, the overall dynamics
still preserve the correct physical trends.
Note that the samples where the model underperforms are characterized by very large displacements and represent outliers
within the data distribution. Hence, a larger error should be expected for these instances. Overall, the test-set NRMSE confirms that this is the most challenging benchmark among the ones presented in this work. Specifically, the x, y, and z displacement components are predicted with an average error of $4.8 \times 10^{-2}$, $4.7 \times 10^{-2}$, and $4.5 \times 10^{-2}$, respectively.

Finally, to investigate the latent space structure, the latent states along the trajectories are projected onto their principal components. Subsequently, the effect of each
principal component is assessed by varying only one of them at a time while keeping the other components equal to their
per-component means. These codes are then decoded to recover the corresponding beam deformation,
as shown in Figure~\ref{fig:effect-pc-beam-cavia-section},
where the deformed free extremity is depicted for selected intermediate interpolation steps.
Interestingly, also in this case, the latent space learned by the Meta-IC-LDNet is physically interpretable.
Indeed, the variation of each latent-state principal component mainly corresponds to a specific mechanical response of the beam,
such as stretching, twisting, bending, or flexural-torsional coupling.

%% file: Sections/conclusions.tex
The Latent Dynamics Network learns the solution operator of time-dependent PDEs, but its original formulation assumes a fixed initial condition, limiting its applicability to real-world settings where the system evolves from varying, unseen initial states. 
This work addresses this limitation by formulating latent-state initialization as an encoder-free adaptation problem, and compares two strategies for solving it, auto-decoding and meta-learning. 
The comparison reveals that the effect of meta-learning extends beyond inference speed: it regularizes the latent-state identification problem, turning an ill-conditioned, multi-modal optimization landscape into a smooth, well-conditioned one. Additionally, meta-learning is consistently associated with latent representations whose topology reflects that of the underlying physical space. 
More broadly, these results suggest that latent representation, state initialization, and latent evolution should not necessarily be designed as separate components. Training the representation for rapid state inference may itself provide a useful inductive bias for learning dynamically meaningful reduced coordinates.

Because the proposed IC-LDNet remains entirely encoder-free, it preserves the resolution independence of the original framework and integrates seamlessly with neural-field-based decoders, enabling training on highly subsampled spatial domains while accurately recovering high-fidelity fields at inference. 
Its versatility was demonstrated across distinct physical domains, from advection-diffusion-reaction and fluid dynamics to nonlinear solid mechanics. Moreover, its rapid inference was further leveraged for an optimal-control problem in fluid dynamics, a many-query task that would be computationally challenging with traditional numerical methods.

Some limitations of this work should be acknowledged. The training of the Meta-IC-LDNet can be sensitive to the choice of hyperparameters and, in some applications, it requires a regularization strategy, chosen on a case-by-case basis, to reach a stable, well-conditioned latent space. Moreover, as observed for the nonlinear-beam test case, predictive accuracy degrades when dealing with outliers with respect to the training distribution. Additionally, in the static- and rotating-cylinder test cases, the validation and test sets contain only 6 and 7 samples each, respectively, which limits the statistical significance of the reported generalization metrics. Finally, the proposed approach was not benchmarked against alternative surrogate models from the literature, as this work focuses on the comparison between encoder-free initialization strategies within the LDNet framework.

Future work will explore the application of the Meta-IC-LDNet to further optimization and control problems, a natural direction given its inference speed. 
It will also be worth investigating whether explicit physical or topological constraints on the latent space can further facilitate training and enforce desired representation properties, building on the inductive bias already observed to emerge from meta-learning itself.

%% file: Sections/appendix_hyperparameters.tex
Here, for each test case presented in this work, we report the selected configurations of hyperparameters, resulting from tuning experiments, as described in Section~\ref{sec:results}. 
\begin{table}[ht!]
    \centering
    \renewcommand{\arraystretch}{1.0}
    \footnotesize
    \begin{tabular}{lcc}
        \toprule
        \multicolumn{3}{c}{\textbf{Test case: ADR}} \\ 
        \midrule
        \textbf{Parameter} & \textbf{AD-IC-LDNet} & \textbf{Meta-IC-LDNet} \\
        \midrule
        \multicolumn{3}{c}{\textit{Architecture}} \\
        \midrule
        Latent dimension $d_s$ & \multicolumn{2}{c}{2} \\
        $\mathcal{NN}_{\mathrm{dyn}}$ architecture (activ.) & FCNN (tanh) & FCNN (SiLU) \\
        $\mathcal{NN}_{\mathrm{dyn}}$ hidden layers (width) & \multicolumn{2}{c}{2 (9)} \\
        $\mathcal{NN}_{\mathrm{rec}}$ architecture (activ.) & FCNN (tanh) & FCNN (SiLU) \\
        $\mathcal{NN}_{\mathrm{rec}}$ hidden layers (width) & \multicolumn{2}{c}{2 (11)} \\
        \midrule
        \multicolumn{3}{c}{\textit{Model training}} \\
        \midrule
        Loss function & \multicolumn{2}{c}{MSE} \\
        Mini-batch size & \multicolumn{2}{c}{800 (full batch)} \\
        Optimizer & \multicolumn{2}{c}{Adam} \\
        Learning rate & \multicolumn{2}{c}{$10^{-4}$} \\
        Validation frequency & \multicolumn{2}{c}{$10^3$ epochs} \\
        Latent-space regularization & \multicolumn{2}{c}{None} \\
        \midrule
        \multicolumn{3}{c}{\textit{Latent-code optimization}} \\
        \midrule
        $\mathcal{T}_{\mathrm{obs}}$ & \multicolumn{2}{c}{Initial condition only} \\
        Outer-loss time domain & - & Remainder of the trajectory \\
        Optimizer & Adam & Gradient Descent \\
        Optimization steps & $10^{3}$ (Val), $10^{4}$ (Test) & 3 \\
        Learning rate & $10^{-2}$ (Val), $10^{-3}$ (Test) & $10^{-2}$ \\
        \bottomrule
    \end{tabular}
    \caption{ADR test case: hyperparameters (AD-IC-LDNet and Meta-IC-LDNet).}
    \label{tab:hyperparameters-comparison-adr}
\end{table}

\begin{table}[b!]
    \centering
    \renewcommand{\arraystretch}{1.0}
    \footnotesize
    \begin{tabular}{lcc}
        \toprule
        \multicolumn{3}{c}{\textbf{Test case: static cylinder}} \\ 
        \midrule
        \textbf{Parameter} & \textbf{AD-IC-LDNet} & \textbf{Meta-IC-LDNet} \\
        \midrule
        \multicolumn{3}{c}{\textit{Architecture}} \\
        \midrule
        Latent dimension $d_s$ & \multicolumn{2}{c}{2} \\
        $\mathcal{NN}_{\mathrm{dyn}}$ architecture (activ.) & \multicolumn{2}{c}{FCNN (SiLU)} \\
        $\mathcal{NN}_{\mathrm{dyn}}$ hidden layers (width) & \multicolumn{2}{c}{2 (16)} \\
        $\mathcal{NN}_{\mathrm{rec}}$ architecture (activ.) & \multicolumn{2}{c}{FiLM-modulated SIREN (sin)} \\
        $\mathcal{NN}_{\mathrm{rec}}$ hidden layers (width) & \multicolumn{2}{c}{2 (128)} \\
        \midrule
        \multicolumn{3}{c}{\textit{Model training}} \\
        \midrule
        Loss function & \multicolumn{2}{c}{MSE} \\
        Mini-batch size & \multicolumn{2}{c}{24} \\
        Optimizer & \multicolumn{2}{c}{Adam} \\
        Learning rate & \multicolumn{2}{c}{$10^{-4}$} \\
        Validation frequency & \multicolumn{2}{c}{25 epochs} \\
        Latent-space regularization & None & \makecell{Curriculum learning \\ (first 2 snaps for 500 ep., +1 every 100 ep.)} \\
        \midrule
        \multicolumn{3}{c}{\textit{Latent-code optimization}} \\
        \midrule
        $\mathcal{T}_{\mathrm{obs}}$ & \multicolumn{2}{c}{Initial condition only} \\
        Outer-loss time domain & - & Entire trajectory \\
        Optimizer & Adam & Gradient Descent \\
        Optimization steps & $10^{3}$ (Val), $10^{4}$ (Test) & 3 \\
        Learning rate & $10^{-2}$ (Val), $10^{-3}$ (Test) & $10^{-2}$ \\
        \bottomrule
    \end{tabular}
    \caption{Static-cylinder test case: hyperparameters (AD-IC-LDNet and Meta-IC-LDNet).}
    \label{tab:hyperparameters-comparison-cyl}
\end{table}

\begin{table}[t!]
    \centering
    \renewcommand{\arraystretch}{1.}
    \footnotesize
    \begin{tabular}{lc}
        \toprule
        \multicolumn{2}{c}{\textbf{Test case: rotating cylinder}} \\ 
        \midrule
        \textbf{Parameter} & \textbf{Meta-IC-LDNet} \\
        \midrule
        \multicolumn{2}{c}{\textit{Architecture}} \\
        \midrule
        Latent dimension $d_s$ & 4 \\
        $\mathcal{NN}_{\mathrm{dyn}}$ architecture (activ.) & FCNN (SiLU) \\
        $\mathcal{NN}_{\mathrm{dyn}}$ hidden layers (width) & 2 (32) \\
        $\mathcal{NN}_{\mathrm{rec}}$ architecture (activ.) & FiLM-modulated SIREN (sin) \\
        $\mathcal{NN}_{\mathrm{rec}}$ hidden layers (width) & 2 (128) \\
        \midrule
        \multicolumn{2}{c}{\textit{Model training}} \\
        \midrule
        Loss function & MSE \\
        Mini-batch size & 14 \\
        Random spatial subsampling & 1250 points (out of $80\,000$) \\
        Fine-tuning points & $54\,500$ \\
        Optimizer & Adam \\
        Learning rate & $10^{-4}$ \\
        Latent-space regularization & \makecell{Curriculum learning \\ (first 3 snaps for $10^3$ ep., +1 every 50 ep.)} \\
        \midrule
        \multicolumn{2}{c}{\textit{Latent-code optimization}} \\
        \midrule
        $\mathcal{T}_{\mathrm{obs}}$ & First 3 snapshots \\
        Outer-loss time domain & Entire trajectory \\
        Optimizer & Gradient Descent \\
        Optimization steps & 3 \\
        Learning rate & $10^{-2}$ \\
        \bottomrule
    \end{tabular}
    \caption{Rotating-cylinder test case: hyperparameters (Meta-IC-LDNet).}
    \label{tab:hyperparameters-rotating-cylinder}
\end{table}

\begin{table}[b!]
    \centering
    \renewcommand{\arraystretch}{1.}
    \footnotesize
    \begin{tabular}{lc}
        \toprule
        \multicolumn{2}{c}{\textbf{Test case: nonlinear beam}} \\ 
        \midrule
        \textbf{Parameter} & \textbf{Meta-IC-LDNet} \\
        \midrule
        \multicolumn{2}{c}{\textit{Architecture}} \\
        \midrule
        Latent dimension $d_s$ & 12 \\
        $\mathcal{NN}_{\mathrm{dyn}}$ architecture (activ.) & FCNN (SiLU) \\
        $\mathcal{NN}_{\mathrm{dyn}}$ hidden layers (width) & 2 (32) \\
        $\mathcal{NN}_{\mathrm{rec}}$ architecture (activ.) & FCNN (SiLU) \\
        $\mathcal{NN}_{\mathrm{rec}}$ hidden layers (width) & 4 (64) \\
        \midrule
        \multicolumn{2}{c}{\textit{Model training}} \\
        \midrule
        Loss function & MSE \\
        Mini-batch size & 24 \\
        Random spatial subsampling & 500 points (out of 1025) \\
        Optimizer & Adam \\
        Learning rate & $2 \times 10^{-4}$ \\
        Latent-space regularization & \makecell{Latent-state penalty \\ (3 additional inner loops at equidistant times)} \\
        \midrule
        \multicolumn{2}{c}{\textit{Latent-code optimization}} \\
        \midrule
        $\mathcal{T}_{\mathrm{obs}}$ & First 3 snapshots \\
        Outer-loss time domain & Entire trajectory \\
        Optimizer & Gradient Descent \\
        Optimization steps & 3 \\
        Learning rate & $10^{-2}$ \\
        \bottomrule
    \end{tabular}
    \caption{Nonlinear-beam test case: hyperparameters (Meta-IC-LDNet).}
    \label{tab:hyperparameters-nonlinear-beam}
\end{table}

%% file: Sections/appendix_training_validation.tex
For each test case, this section presents the training and validation curves for the final hyperparameter configurations described in Appendix~\ref{sec:appendix-hyper}. For the AD-IC-LDNet we also report the MSE evolution during the optimization of the initial codes for the unseen test-set examples. For the rotating-cylinder test case, we also report the fine-tuning run.
In all figures, the top axis reports the elapsed time.

\begin{figure}[ht!]
    \centering
    \begin{subfigure}{0.49\textwidth}
        \centering
        \includegraphics[scale=0.85]{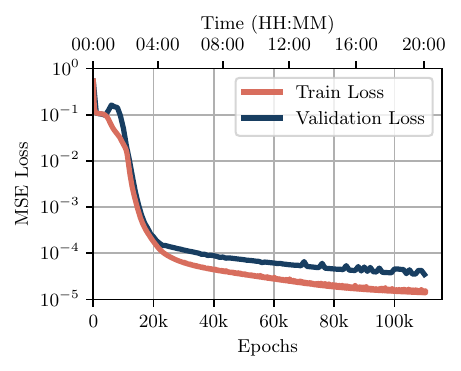}
        \caption{Training and validation losses against epochs.}
        \label{subfig:train_val_adr_autodec}
    \end{subfigure}
    \begin{subfigure}{0.49\textwidth}
        \centering
        \includegraphics[scale=0.85]{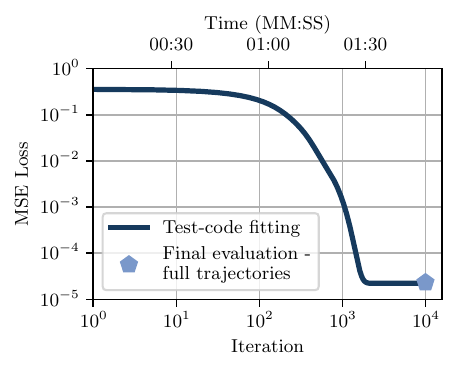}
        \caption{Test-set latent-code optimization.}
        \label{subfig:fit_latent_codes_adr_autodec}
    \end{subfigure}
    \caption{ADR test case: training, validation and testing (AD-IC-LDNet).}
    \label{fig:train_val_test_adr_autodec}
    \end{figure}

\begin{figure}[ht!]
    \centering
    \includegraphics[scale=0.85]{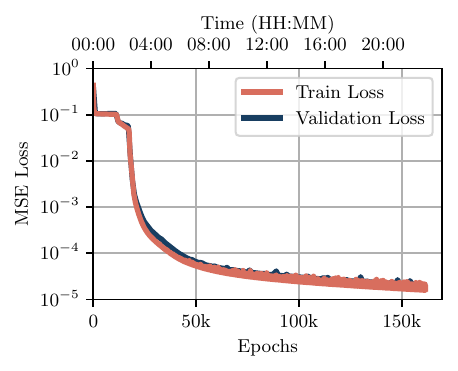}
    \caption{ADR test case: training and validation losses against epochs (Meta-IC-LDNet).}
    \label{fig:train_val_adr_cavia}
\end{figure}

\begin{figure}[ht!]
    \centering
    \begin{subfigure}{0.49\textwidth}
        \centering
        \includegraphics[scale=0.85]{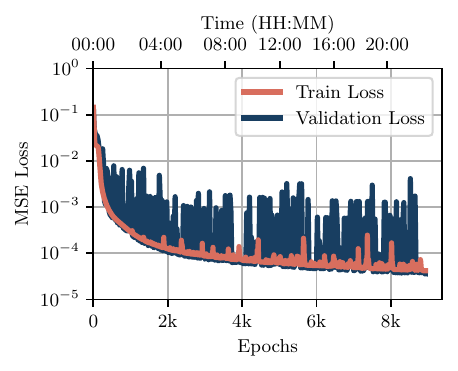}
        \caption{Training and validation losses against epochs.}
        \label{subfig:train_val_cyl100_autodec}
    \end{subfigure}
    \begin{subfigure}{0.49\textwidth}
        \centering
        \includegraphics[scale=0.85]{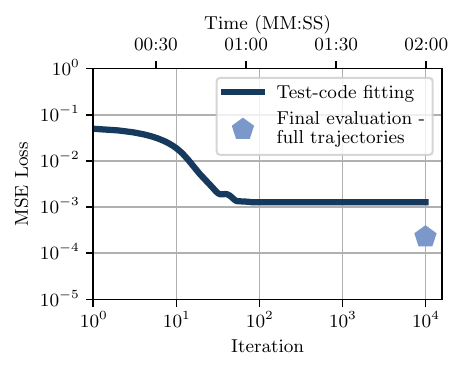}
        \caption{Test-set latent-code optimization.}
        \label{subfig:fit_latent_codes_cyl100_autodec}
    \end{subfigure}
    \caption{Static-cylinder test case: training, validation and testing (AD-IC-LDNet).}
    \label{fig:train_val_test_cyl100_autodec}
\end{figure}

\begin{figure}[ht!]
    \centering
    \includegraphics[scale=0.85]{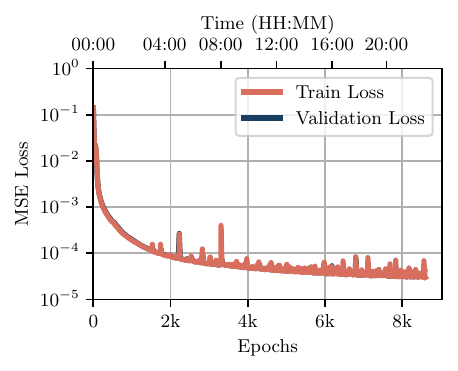}
    \caption{Static-cylinder test case: training and validation losses against epochs (Meta-IC-LDNet).}
    \label{fig:train_val_cylinder_cavia}
\end{figure}

\begin{figure}[ht!]
    \centering
    \begin{subfigure}{0.49\textwidth}
        \centering
        \includegraphics[scale=0.85]{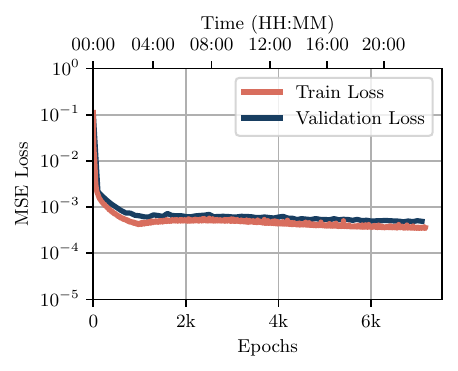}
    \caption{Training and validation losses against epochs.}
    \label{subfig:train_val_randomrot_cavia}
    \end{subfigure}
    \begin{subfigure}{0.49\textwidth}
        \centering
        \includegraphics[scale=0.85]{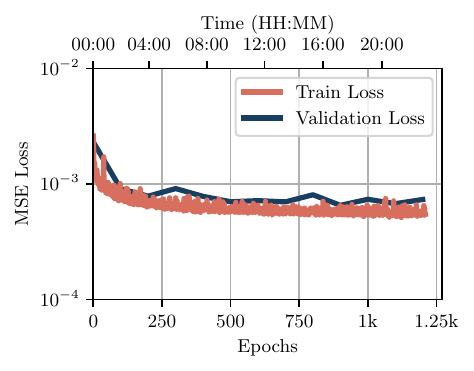}
        \caption{Fine-tuning on a denser grid of points.}
        \label{subfig:train_val_finetune}
    \end{subfigure}
    \caption{Rotating-cylinder test case: training and fine-tuning (Meta-IC-LDNet).}
    \label{fig:train_ft_randomrot_cavia}
\end{figure}

\begin{figure}[ht!]
    \centering
    \includegraphics[scale=0.85]{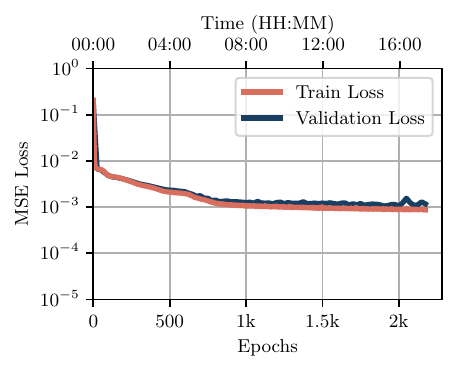}
    \caption{Nonlinear-beam test case: training and validation losses against epochs (Meta-IC-LDNet).}
    \label{fig:train_val_beam_cavia}
\end{figure}